\documentclass[journal]{IEEEtran} 
\usepackage{graphicx,psfrag,amsmath,amsfonts,verbatim,url,booktabs,array,float}
\usepackage[small,bf]{caption}
\usepackage[short]{optidef}
\usepackage{algorithm}
\usepackage[noend]{algpseudocode}
\usepackage{caption,subcaption}
\usepackage{censor}
\usepackage{graphicx}
\usepackage{censor}
\usepackage{multirow}
\usepackage{lipsum}
\usepackage{soul}
\usepackage{enumitem}

\usepackage{hyperref}
\usepackage{caption} 

\usepackage{xcolor}
\definecolor{orange}{RGB}{255,127,0}
\definecolor{cyan}{RGB}{0,255,255}
\definecolor{magenta}{RGB}{255,0,255}

\usepackage{pgfplots}
\pgfplotsset{
    legend image with text/.style={
        legend image code/.code={%
            \node[anchor=center] at (0.3cm,0cm) {#1};
        }
    },
}
\usepackage{tikz}
\usepackage{tikzscale}
\usetikzlibrary{matrix}
\usetikzlibrary[pgfplots.groupplots]
\usetikzlibrary{arrows.meta}

\usetikzlibrary{backgrounds}
\usepackage{pgfplots}
\usepgfplotslibrary{groupplots}

\usepackage[numbers]{natbib}
\usepackage{multicol}

\usepackage{wrapfig}

\usepackage{xcolor}

\newcommand{\mac}[1]{\textcolor{black}{#1}}

\newcommand{\js}[1]{\textcolor{black}{#1}}
\newcommand{\qc}[1]{{\color{black}{#1}}}

\title{Dojo: A Differentiable Physics Engine for Robotics}

\author{Taylor A. Howell$^{1*}$, Simon Le Cleac'h$^{1*}$, Jan Br\"udigam$^{2}$, Qianzhong Chen$^{1}$, Jiankai Sun$^{1}$,\\ J. Zico Kolter$^{3}$, Mac Schwager$^{1}$, and Zachary Manchester$^{4}$%
    \thanks{$^{1}$ Stanford University, Stanford, CA 94305, USA.
		{\tt\footnotesize \{thowell, simonlc, qchen23, jksun, schwager\}@stanford.edu}}%
    \thanks{$^{2}$ School of Computation, Information and Technology, Technical University of Munich, Munich, 80333, Germany.
		{\tt\footnotesize jan.bruedigam@tum.de}}%
    \thanks{$^{3}$ Department of Computer Science, Carnegie Mellon University, Pittsburgh, PA 15213, USA.
		{\tt\footnotesize zkolter@cs.cmu.edu}}%
    \thanks{$^{4}$ The Robotics Institute, Carnegie Mellon University, Pittsburgh, PA 15213, USA.
		{\tt\footnotesize zacm@cmu.edu}}
    \thanks{\textit{(Corresponding author: S. Le Cleac'h)}}
    \thanks{$^{*}$ These authors contributed equally to this work.}
}

\begin{document}
\maketitle

\begin{abstract}
    We present Dojo, a differentiable physics engine for robotics that prioritizes stable simulation, accurate contact physics, and differentiability with respect to states, actions, and system parameters. \qc{Dojo models hard contact and friction with a nonlinear complementarity problem with second-order cone constraints.  We introduce a custom primal-dual interior-point method to solve the second order cone program for stable forward simulation over a broad range of sample rates.  We obtain smooth gradient approximations with this solver through the implicit function theorem, giving gradients that are useful for downstream trajectory optimization, policy optimization, and system identification applications. Specifically, we propose to use the central path parameter threshold in the interior point solver as a user-tunable design parameter.  A high value gives a smooth approximation to contact dynamics with smooth gradients for optimization and learning, while a low value gives precise simulation rollouts with hard contact.  We demonstrate Dojo's differentiability in trajectory optimization, policy learning, and system identification examples. We also benchmark Dojo against MuJoCo, PyBullet, Drake, and Brax on a variety of robot models, and study the stability and simulation quality over a range of sample frequencies and accuracy tolerances.  Finally, we evaluate the sim-to-real gap in hardware experiments with a Ufactory xArm 6 robot. Dojo is an open source project implemented in Julia with Python bindings, with code available at \url{https://github.com/dojo-sim/Dojo.jl}.}
\end{abstract}


\section{Introduction}
    The last decade has seen immense advances in learning-based methods for policy optimization and trajectory optimization in robotics, e.g., for dexterous manipulation \cite{andrychowicz2020learning,akkaya2019solving}, quadrupedal locomotion \cite{lee2020learning, kumar2021rma}, and pixels-to-torques control \cite{levine2018learning}.  These advances have largely hinged on innovations in learning architectures, large scale optimization algorithms, and large datasets. In contrast, there has been comparatively little work on the lowest level of the robotics reinforcement learning stack: the \textit{physics engine}. \qc{We argue that core improvements in physics engines can enable future advancements in robotics, and we present Dojo as a physics engine that embodies several such advances.} 
	
	Physics engines that simulate rigid-body dynamics with contact are utilized for trajectory optimization, reinforcement learning, system identification, and dataset generation for domains ranging from locomotion to manipulation. To overcome the sim-to-real gap \cite{zhao2020sim} and to be of practical value in real-world applications, an engine should provide stable simulation, accurately reproduce a robot's dynamics, and ideally, be differentiable to enable the use of efficient gradient-based optimization methods.
    
    \begin{figure}[t]
        \centering
        \includegraphics[height=5.0cm]{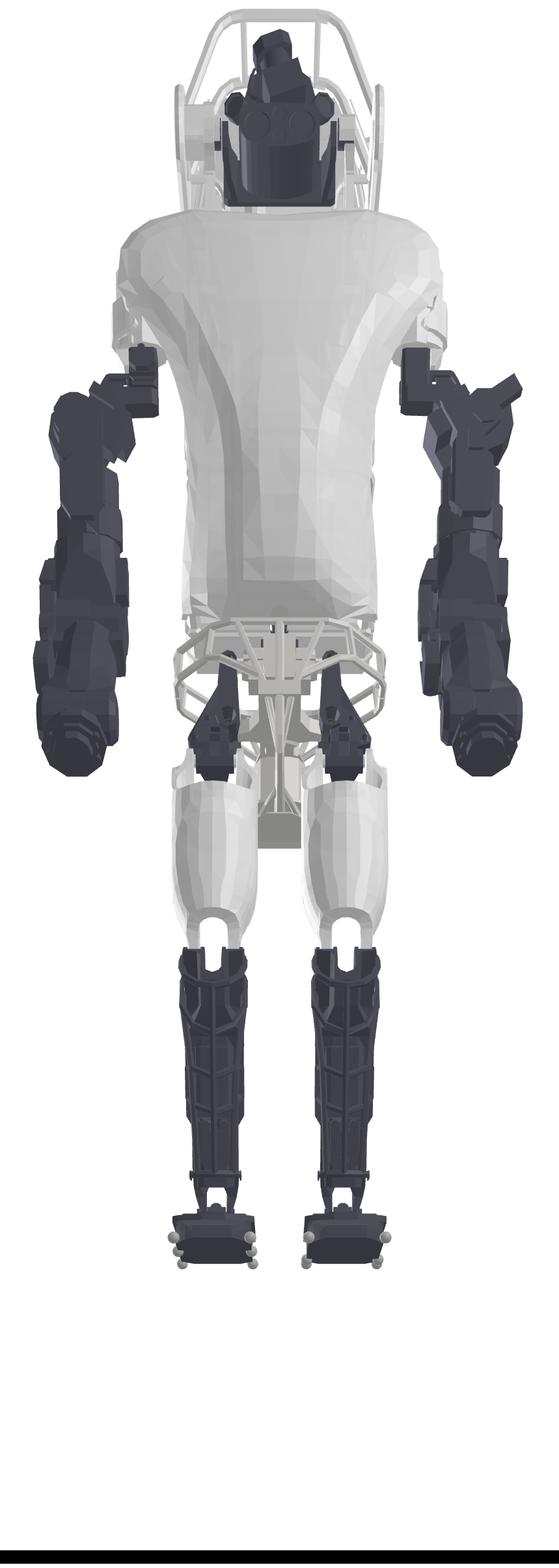} 
        \includegraphics[height=5.0cm]{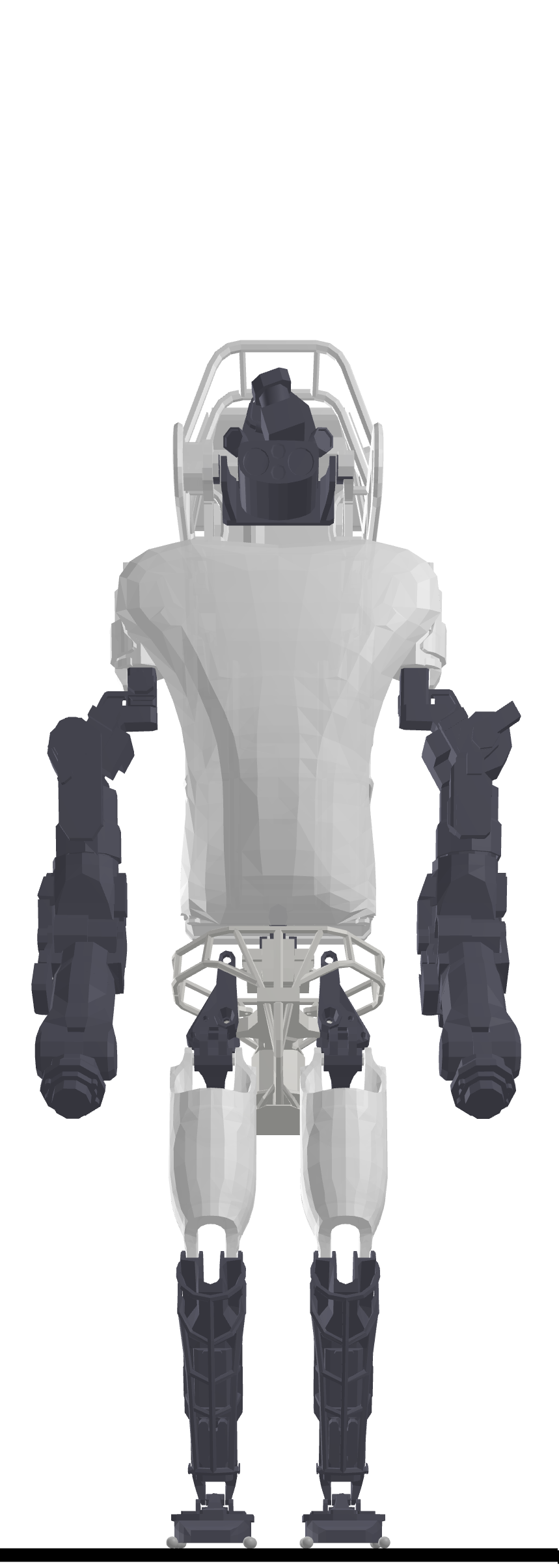}
        \includegraphics[height=5.0cm]{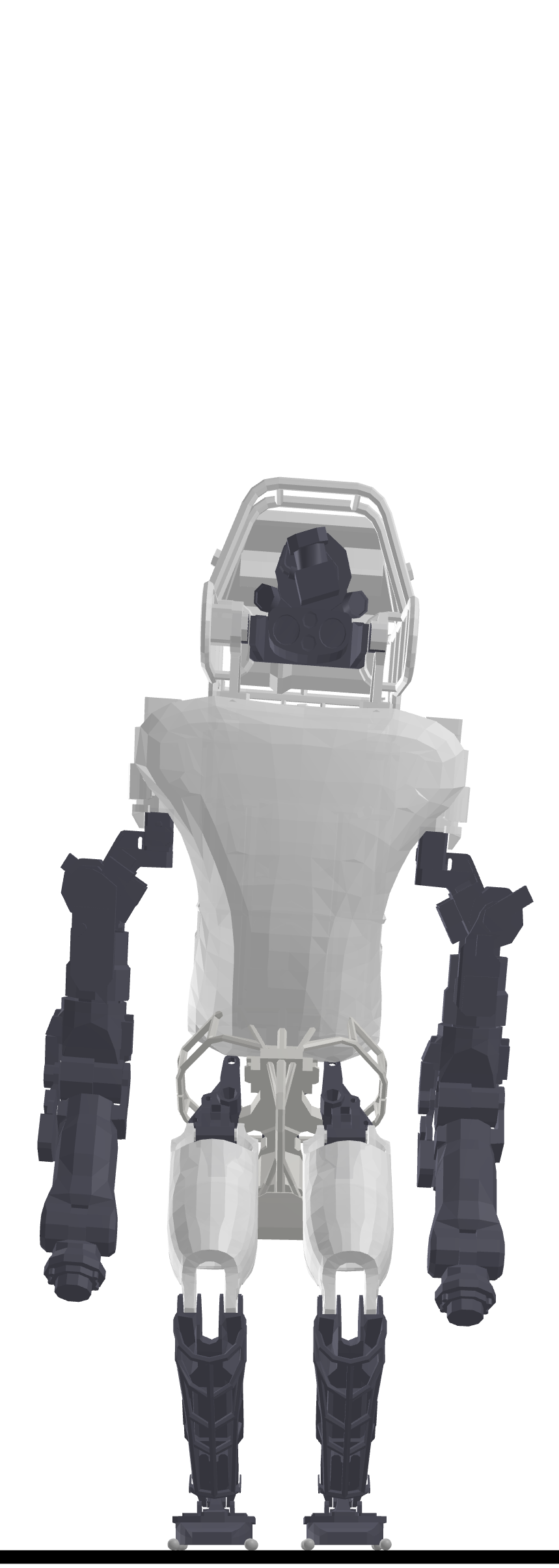}
        \includegraphics[height=5.0cm]{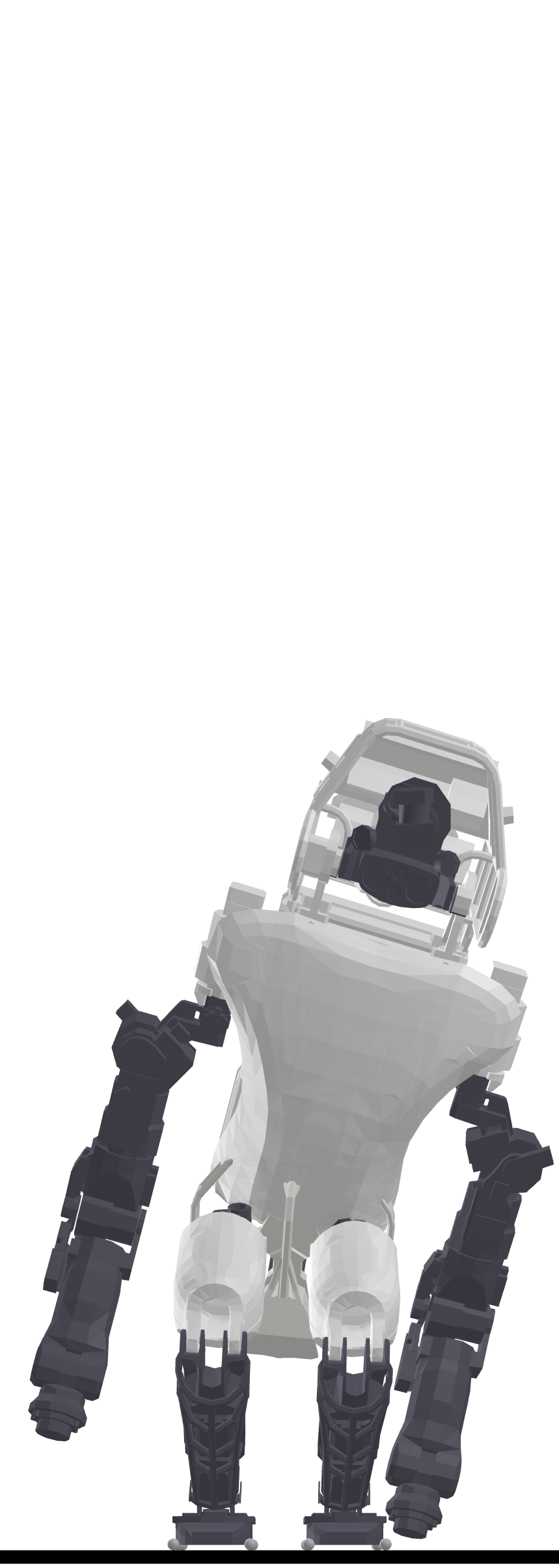}
        \caption{Atlas drop simulation. Dojo simulates this system with 403 maximal-coordinates states, 30 joint constraints, 36 inputs, and 8 contact points in real-time at 65 Hz. Dojo respects floor-feet penetration constraints to machine precision.  \qc{Other simulators struggle to maintain the floor contact constraint, especially at low simulation rates.}}
        \label{atlas_drop}
    \end{figure}
    
	In recent years, a number of physics engines \cite{drake, freeman2021brax, werling2021fast, geilinger2020add, hu2019difftaichi, heiden2020neuralsim,nvidia_isaac_sim,Genesis}
    have been developed and utilized for robotics. \qc{These works have advanced the state of the art in robot simulation, particularly offering differentiability \cite{freeman2021brax,hu2019difftaichi,werling2021fast, geilinger2020add}, multi-physics simulation \cite{nvidia_isaac_sim,Genesis,hu2019difftaichi,macklin2014unified}, and the incorporation of learnable dynamics residuals \cite{heiden2020neuralsim}. Recent work is also moving towards a convergence of fully learning based world models and traditional physics-based simulators \cite{agarwal2025cosmos,zhou2024dino}, representing an exciting new frontier in robot simulation.} 
    
    \qc{In this work, we propose to further advance the state of the art by focusing on the underlying numerics of the physics engine, introducing features that can be adopted throughout the existing simulation ecosystem to improve performance in a variety of ways.  We package these numerical improvements in a new simulation engine, Dojo, to highlight their favorable properties over exiting numerical techniques commonly used in physics simulators.  
    
    Specifically, we introduce two new contributions specific to Dojo: (i) We propose a custom primal-dual interior point solver for stably solving the complementary problem for forward simulation.  This solver allows for low sample rates while maintaining stable and accurate contact simulation, thereby alleviating the \emph{vanishing/exploding} gradient problem that appears when differentiating through high sample rate rollouts common in other differentiable simulators.  (ii) We obtain tunable gradient information through implicit differentiation of the interior point solver, obtaining user-defined smoothness of the gradients through contact events by tuning the central path parameter.  This gives informative gradients through contact events for policy and trajectory optimization (with a high central path parameter), while also enabling sharp, physically accurate simulation rollouts (with a low central path parameter).  This is in contrast to existing methods that either deliver non-smooth gradients that are not informative for trajectory or policy optimization, or expensive sampling-based gradient approximations that require a large number of calls to the engine, leading to slow computation.
    
    Additionally, we also incorporate the following features that have been presented in previous works, but are not typically implemented in existing physics engines: (a) We use a variational integration scheme for strong energy and momentum conservation in non-contact regimes. (b) We use full nonlinear complementarity constraints with nonlinear friction cones to avoid numerical artifacts like interpenetration of rigid bodies (e.g., a robot foot sinking through the floor) and creep (e.g., objects that should be at rest incorrectly sliding) commonly seen in robotics simulators \qc{\cite{parmar2021fundamental}}. (c) We use maximal coordinates, explicitly representing the 6 degree of freedom pose of each rigid link, and the internal constraint forces that bind them together.}
	
	\begin{table*}[t]
		\centering
		\caption{Comparison of physics engines used for robotics. \qc{Several simulators like MuJoCo and Drake have a selection of integrators that users can choose. We demonstrate their default/most-widely-used integrators.}}
		\begin{tabular}{c c c c c c c}
			\toprule
			\textbf{Engine} & \textbf{Application} & \textbf{Integrator} & \textbf{State} & \textbf{\qc{Contact Softness}} & \textbf{\qc{Contact Model/Solver}} & \textbf{Gradients} \\
			\toprule
			MuJoCo & robotics & RK4 & minimal & soft & Newton & finite difference \\
			Drake & robotics & implicit Euler & minimal & soft/hard & \qc{penalty-based/SAP} & randomized smoothing\\
			Bullet & graphics & implicit Euler & minimal & soft/hard & \js{iterative} & \qc{finite difference} \\
			DART & robotics & implicit Euler & minimal & hard & LCP & subgradient\\
		    PhysX & graphics & explicit & minimal & soft & iterative & finite difference \\
		    Brax & graphics & explicit & maximal & soft & iterative & \qc{analytical}\\
		    \hline
			\textbf{Dojo} & \textbf{robotics} & \textbf{variational} & \textbf{maximal} & \textbf{hard} & \textbf{NCP} & \textbf{implicit gradient} \\
			\toprule
		\end{tabular}
		\label{engine_comparison}
	\end{table*}
	
	\qc{The Dojo physics engine is designed around these core numerical innovations with the goal of advancing robot simulation for trajectory optimization and motion planning, control, reinforcement learning, system identification, and for generating high-quality datasets for learning and validation. 
    
To demonstrate the advantages of this combination of numerical features, we benchmark Dojo against MuJoCo \cite{todorov2012mujoco}, Drake \cite{drake}, and Brax \cite{freeman2021brax} for computational speed in simulations with four different robot platforms, showing Dojo delivers a comparable performance as other differentiable simulators.  We demonstrate Dojo's numerical stability and accuracy over sample rates down to 20Hz, allowing for lower sample rates than other simulators, leading to fewer rollout steps, and therefore less severe vanishing/exploding gradients.  We also compare Dojo's primal-dual solver versus a more common primal only interior point solver, showing improved convergence speed and constraint satisfaction. We demonstrate Dojo's differentiability in trajectory optimization, policy optimization, and system identification examples in comparison with sampling-based gradient approximations.  Finally, we show Dojo's low sim-to-real gap in hardware experiments with a UFactory xArm 6.  

In summary, the key contributions of this paper as integrated into the Dojo simulator include:} 
	\begin{enumerate}[label = (\roman*)]
    \item \qc{a custom primal-dual interior-point method for giving stable, accurate simulation over sample rates as low as 20Hz,
    \item analytic gradients through contact efficiently computed via implicit differentiation of the interior-point solver with a user-defined smoothness approximation set by the central path parameter,
	\item incorporation of variational integration and a nonlinear complementarity problem (NCP) model for accurate contact dynamics (previously introduced in literature, but not yet integrated in a simulator).}
	\end{enumerate}

	In the remainder of this paper, we first provide an overview of related state-of-the-art physics engines in Section \ref{sota}. We then summarize important technical background in Section \ref{background}. Next, we present Dojo, and its key features in Section \ref{dojo}. Simulation, planning, policy optimization, and system identification examples, \qc{and hardware sim-to-real gap evaluation} are presented in Section \ref{examples}. Finally, we conclude with a discussion of limitations and future work in Section \ref{conclusion}.

\section{Related Work} \label{sota}
    \js{In this section, we provide an overview of several popular physics engines, focusing on their physical fidelity, underlying optimization algorithms, and capability to compute gradients.}
    \subsection{Mainstream Physics Engines}
    \subsubsection{MuJoCo}
	In the learning community, \emph{MuJoCo}~\cite{todorov2012mujoco} has become a standard for benchmarking reinforcement learning algorithms using the OpenAI Gym environments~\cite{brockman2016openai}.
	MuJoCo utilizes minimal-coordinates representations, and employs both semi-implicit Euler and explicit fourth-order Runge-Kutta integrators to simulate multi-body systems. These integrators often require small time steps, particularly for systems experiencing contact, and typically sample rates of hundreds to thousands of Hertz are required for stable simulation, which a mature and efficient implementation is able to achieve at much faster than real-time rates. However, these high rates can prove a challenge for control tasks, such as reinforcement learning settings where vanishing or exploding gradients are exacerbated over long horizons with many time steps \qc{\cite{suh2022differentiable,parmas2018pipps,metz2021gradients}}.
	
	Impact and friction are modeled using a smooth, convex contact model \cite{todorov2014convex}. While this approach reliably computes contact forces, it introduces unphysical artifacts, and contact forces at a distance (i.e., while not in contact) \qc{\cite{SimBenchmark2025}}, and the default friction model introduces creep and velocity drift during sliding. Additionally, achieving good simulation behavior often requires system-specific tuning of multiple solver parameters. Further, the ``soft" contact model is computed using a \textit{primal} optimization method, meaning that as parameters are set to produce ``hard'' or more realistic contact, the underlying optimization problem becomes increasingly ill-conditioned and difficult to solve. \mac{For RL methods, where obtaining a large volume of rollouts quickly is more important than preserving physical fidelity of each rollout, these compromises have been effective.  However, for control and planning with real robot hardware they can be problematic. For example}, it is often not possible to eliminate unphysical artifacts from the simulation to produce realistic results. The lack of smooth gradients is also a major challenge in deep learning \cite{parmar2021fundamental}, where contact dynamics must \js{often be smoothened unrealistically in order} to make learning progress.
    Analytical gradients are not provided by the engine, and instead require finite-difference schemes \cite{tassa2012synthesis} that are computationally expensive.
    This approach requires multiple calls to the engine, which can be expensive if not performed in parallel.
    
\subsubsection{Drake}
    \qc{\emph{Drake} \cite{drake}, designed for robotics applications, prioritizes physical accuracy and flexibility in simulation. Its contact modeling primarily employs a penalty method, where contact forces are approximated using stiff springs, based on the Hunt-Crossley model \cite{hunt1975coefficient}. This approach provides a compliant contact model but requires small time steps to ensure stability, as the stiffness can lead to numerical challenges such as instability and gradient explosion~\cite{suh2022differentiable}. Recently, Drake has introduced the soft articulated-body dynamics with compliant contact (SAP) method~\cite{castro2022unconstrained}, which formulates contact as a time-stepping optimization problem. SAP introduces compliance to relax the contact model, making it robust for simulating articulated systems, similar in spirit to methods used in MuJoCo~\cite{todorov2012mujoco}. Drake offers flexibility in integrator choices, including advanced error-controlled methods, ensuring stable and accurate simulation for various dynamics. Gradients in Drake are currently computed using Eigen's autodiff framework through the penalty method, but this approach is less ideal for stiff systems. While SAP could theoretically utilize the implicit function theorem for gradient computation, this feature has not been implemented yet. Additionally, methods like randomized smoothing for returning gradients~\cite{suh2022bundled} provide alternative strategies for handling contact-rich dynamics outside of Drake's native capabilities.}
    
\subsubsection{\js{Other Engines}}
    The popular robotics simulator \emph{Gazebo} \cite{koenig2004design} can utilize several different physics engines to simulate multi-body contact dynamics, Bullet \cite{coumans2019} and DART \cite{lee2018dart} are common choices. These engines model hard contact dynamics with an LCP formulation. Automatic differentiation tools have been utilized to compute gradients \cite{heiden2020neuralsim}, \qc{\cite{murthy2020gradsim}, \cite{heiden2021disect}}. However, because of the discontinuous nature of contact dynamics, this approach will return \mac{discontinuous gradients at contact events, which are less useful for gradient based trajectory or policy optimization}. Heuristics have been proposed to enumerate contact modes in order to select informative gradients \cite{werling2021fast}. However, this approach scales poorly with the number of contact mode switches.

    Engines designed for hardware accelerators (e.g., GPUs), including \emph{Brax} \cite{freeman2021brax} and \emph{PhysX} \cite{physx2022engine}, typically utilize simplified contact dynamics. Additionally, these engines usually require system-specific tuning and their simulation results typically prioritize speed over physical fidelity, so a large number of rollouts can be obtained quickly to train learning based policies.

\js{\subsection{Differentiable Physics Engines} 
In contrast to traditional physics engines, differentiable physics engines present promising opportunities for robotics by incorporating physical models into auto-differentiation frameworks~\cite{newbury2024review}. Prior research has explored differentiation across various domains, such as contact and friction models~\cite{toussaint2018differentiable, de2018end, song2020learning, song2020identifying, degrave2019differentiable, wu2017learning}, latent state models~\cite{guen2020disentangling, schenck2018spnets, jaques2019physics, heiden2019interactive}, volumetric soft bodies~\cite{hu2019chainqueen, liang2019differentiable, hu2019difftaichi}, and particle dynamics~\cite{schenck2018spnets, li2018learning}. Additionally, system identification using parameterized physics models~\cite{salzmann2011physically, brubaker2010physics, kozlowski2012modelling, wensing2017linear, brubaker2009estimating, bhat2003estimating, liu2005learning, grzeszczuk1998neuroanimator, sutanto2020encoding, wang2020first} and inverse simulation techniques~\cite{murray2000inverse} have also been explored.
Furthermore, there has been considerable research on smooth gradient computation~\cite{kim2022contact}, which aids in reducing noise in gradient-based model explanations and ensures stable gradient calculations. Examples of such systems include Warp~\cite{macklin2022warp} and Brax~\cite{freeman2021brax}, both of which utilize the XPBD~\cite{macklin2016xpbd} model for contact simulation. The GradSim framework~\cite{jatavallabhula2021gradsim} combines a physics simulator with a differentiable rendering pipeline. Key components of these simulators include gradient calculation, dynamics models, contact models, and integrators. These simulators are capable of leveraging gradient-based optimization techniques to enhance real-to-sim transfer capabilities.}

\qc{Gradients of rigid body dynamics can be useful in robotics for various purposes. Applications include system identification \cite{carpentier2018analytical},  controller design \cite{giftthaler2017automatic}, controller tuning \cite{cheng2024difftune}, trajectory optimization \cite{chen2023simultaneous}, and policy optimization \cite{mora2021pods}. However, gradients are typically computed through autodifferentiating through a simulation rollout by expressing the rollout as a computation graph, making use of mature auto-diff capabilities, e.g. in PyTorch.  Unfortunately, this approach often leads to numerical instability as a long chain of differentiation tends toward infinity or zero as determined by the eigenvalues of the Jacobians in the computation graph: the so called vanishing/exploding gradients problem.}
    
\qc{Dojo uses an optimization solver to propagate a trajectory in time, making it impossible to apply back propagation to compute gradients.  Instead, we use implicit differentiation, which has been explored recently to obtain gradients of the solution to optimization problems with respect to problem parameters~\cite{de2018end,le2021differentiable}. Implicit gradients are computed using the implicit function theorem, rather than directly auto-diffing through through a simulation rollout, in hopes of producing more stable gradient computations.}

The properties and characteristics of several of these existing engines are summarized in Table~\ref{engine_comparison}. We find that none of the existing engines prioritize two of the most important attributes of robotics: physical accuracy and useful differentiability. This motivates our development of Dojo as a physics engine for robotics applications.

Building on prior work \cite{brudigam2021linear}, Dojo utilizes the open-source maximal-coordinates dynamics library \texttt{ConstrainedDynamics.jl} and efficient graph-based linear-system solver \texttt{GraphBasedSystems.jl}. However, unlike this previous work, Dojo has an improved contact model, specifically with regard to friction; and utilizes a more efficient, reliable, and versatile interior-point solver for the NCP.

\section{Background} \label{background}
	
	Here we review existing methods of \qc{complementarity-based contact models, implicit differentiation, maximal coordinates, and variational integrators that are used in Dojo.}

    \subsection{Complementarity-Based Contact Models} 
    Impacts and friction can be modeled through constraints on the system's configuration and the applied contact impulses.

	\emph{Impact:} For a system with $P$ contact points, we define a signed-distance function, $\phi : \mathbf{Z} \rightarrow \mathbf{R}^P$, subject to the following element-wise constraint
	\begin{equation} 
	    \phi(z) \geq 0. \label{sdf}
	\end{equation}
	\qc{Where $z$ represents the system configuration (i.e., the pose of robot and other objects in the simulation environment).} Impact forces with magnitude $\gamma \in \mathbf{R}^P$ are applied to the bodies' contact points in the direction of their surface normals in order to enforce \eqref{sdf} and prevent interpenetration. \qc{Collision points\footnote{\qc{Collision geometries are currently limited to simple shape primitives (e.g., four points on the foot of a humanoid instead of using a full mesh). While this is a limitation of the current implementation, it is common in other simulators to manually edit contact geometries for critical parts of the robot, such as the feet of a humanoid or quadruped.}} are checked for constraint satisfaction at each iteration of our  primal-dual interior-point solver that will be introduced in \ref{sec:PD-IP_solver}.} A non-negative constraint
	\begin{align}
		\gamma \geq 0, \label{impact_ineq}
	\end{align}
	enforces physical behavior that impulses are repulsive (e.g., the floor does not attract bodies), and the complementarity condition
	\begin{align}
		\gamma \circ \phi(z) = 0, \label{impact_comp}
	\end{align}
	where $\circ$ is an element-wise product operator, enforces zero force if the body is not in contact and allows non-zero force during contact.
	
	\emph{Friction:} 
	Coulomb friction instantaneously maximizes the dissipation of kinetic energy between two objects in contact. For a single contact point, this physical phenomenon can be modeled by the following optimization problem:
	\begin{equation}
		\begin{array}{ll}
		\underset{b}{\mbox{minimize}} & v^T b\\
		\mbox{subject to} &\|b\|_2\leq c_{\mathrm{f}} \gamma, \\
		\end{array} \label{mdp}
	\end{equation}
	where $v \in \mathbf{R}^{2}$ is the tangential velocity at the contact point, $b \in \mathbf{R}^2$ is the friction force, and $c_{\mathrm{f}} \in \mathbf{R}_{+}$ is the coefficient of friction between the two objects \cite{moreau2011unilateral}. 

    \begin{figure}[t]
		\centering
		\includegraphics[width=0.48\columnwidth]{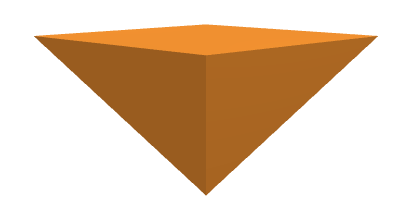}
		\includegraphics[width=0.48\columnwidth]{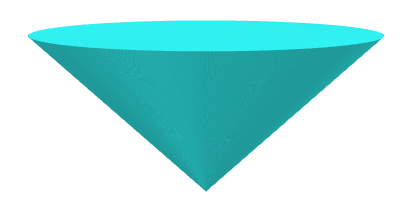}
		\hfill
		\caption{Friction-cone comparison. Linearized double-parameterized (left) and nonlinear second-order (right) cones.}
		\label{friction_cones}
	\end{figure}
 
	This problem is naturally a convex second-order cone program, and can be efficiently and reliably solved \cite{lobo1998applications}. Classically, other works solve an approximate version of \eqref{mdp},
	\begin{equation}
		\begin{array}{ll}
		\underset{\beta}{\mbox{minimize}} & \begin{bmatrix} v^T & -v^T\end{bmatrix} \beta \\
		\mbox{subject to} & \beta^T \textbf{1} \leq c_{\mathrm{f}} \gamma, \\
		& \beta \geq 0, \\
		\end{array} \label{mdp_linear}
	\end{equation}
	which satisfies the LCP formulation. Here, the friction cone is linearized (Fig. \ref{friction_cones}) and the friction vector, $\beta \in \mathbf{R}^{4}$, is correspondingly overparameterized and subject to additional positivity constraints \cite{stewart1996implicit}. 
	
	The optimality conditions of \eqref{mdp_linear} and constraints used in the LCP are
	\begin{align}
	& \begin{bmatrix} v^T & -v^T\end{bmatrix}^T + \psi \mathbf{1} - \eta = 0, \label{mdp_eq}\\
	& c_{\mathrm{f}} \gamma -\beta^T \textbf{1} \geq 0, \label{mdp_cone}\\
	& \psi \cdot (c_{\mathrm{f}} \gamma - \beta^T \textbf{1}) = 0, \label{mdp_cone_comp}\\
	& \beta \circ \eta = 0, \label{mdp_friction_comp}\\
	& \beta, \psi, \eta \geq 0, \label{mdp_ineq}
	\end{align}
	where $\psi \in \mathbf{R}$ and $\eta \in \mathbf{R}^{4}$ are the dual variables associated with the friction cone and positivity constraints, respectively. \qc{In practice, $\psi$ acts as a flag to judge if the object has relative movement to the ground, while $\eta$ is used to enforce the constraint that friction forces align with the vertices of the linearized friction cone approximation,} and $\textbf{1}$ is a vector of ones. 
	
	The primary drawback of this formulation is that the optimized friction force will naturally align with the vertices of the cone approximation, which may not align with the velocity vector of the contact point. Thus, the friction force does not perfectly oppose the movement of the system at the contact point. Unless the pyramidal approximation is improved with a finer discretization, incurring increased computational cost, unphysical velocity drift will occur. Additionally, the LCP contact model requires a linearized form of the dynamics \eqref{implicit_dynamics} and a linear approximation of the signed-distance functions (\ref{sdf}-\ref{impact_comp}), both of which negatively impact physical accuracy.  \mac{To avoid these problems, in Dojo we solve the full NCP with nonlinear friction cone, and develop a new primal-dual interior point method to reliably solve this more difficult optimization problem.}

	\subsection{Implicit Differentiation}
    An implicit function, $r : \mathbf{R}^{n_w} \times \mathbf{R}^{n_\theta} \rightarrow \mathbf{R}^{n_w}$, is defined as $r(w^*; \theta) = 0$
	for solution $w^* \in \mathbf{R}^{n_w}$ and problem data $\theta \in \mathbf{R}^{n_\theta}$. At a solution point, the sensitivities of the solution with respect to the problem data, i.e., $\partial w^* / \partial \theta$, can be computed using the implicit function theorem \cite{dini1907lezioni}
    \begin{equation}
    	\frac{\partial w^*}{\partial \theta} = -\Big(\frac{\partial r}{\partial w}\Big)^{-1} \frac{\partial r}{\partial \theta}. \label{solution_sensitivity}
    \end{equation}
    Newton's method is typically employed to find solutions $w^*$. When the method succeeds, the sensitivity \eqref{solution_sensitivity} can be computed and the factorization of $\partial r / \partial w$ used to find the solution is reused to efficiently compute sensitivities at very low computational cost, using only back-substitution. Additionally, each element of the sensitivity can be computed in parallel.  Dojo uses implicit method to obtain gradients for simulation rollouts with respect to control inputs, dynamics parameters, or initial conditions.

    \subsection{Maximal-Coordinates State Representation} 
    \label{maximal-coordinates}
	
	Most multi-body physics engines utilize minimal- or joint-coordinate representations for dynamics because of the small number of states and convenience of implementation. This results in small, but dense, systems of equations. In contrast, maximal-coordinates explicitly represent the position, orientation, and velocities of each body in a multi-body system. This produces large, sparse systems of equations that can be efficiently solved, including in the contact setting. We provide an overview, largely based on prior work \cite{brudigam2020linear}, of this representation. 
    
    A single rigid body is defined by its mass and inertia, and has a configuration, $x = (p, q) \in \mathbf{X} = \mathbf{R}^3 \times \mathbf{H}$, comprising a position $p$ and unit quaternion $q$, where $\mathbf{H}$ is the space of four-dimensional unit quaternions. We define the implicit discrete-time dynamics $F : \mathbf{X} \times \mathbf{X} \times \mathbf{X} \rightarrow \mathbf{R}^{6}$ as
	\begin{equation}
	    F(x_{-}, x, x_{+}) = 0, \label{implicit_dynamics}
	\end{equation}
	where we indicate the previous and next time steps with minus ($-$) and plus ($+$) subscripts, respectively, and the current time step without decoration.
	We employ a variational integrator that has desirable energy and momentum conservation properties \cite{marsden2001discrete}. Linear and angular velocities are handled implicitly via finite-difference approximations.
 
	For a two-body system with bodies $a$ and $b$ connected via a joint---common types include revolute, prismatic, and spherical---we introduce a constraint, $k : \mathbf{X} \times \mathbf{X} \rightarrow \mathbf{R}^{l}$, that couples the two bodies 
	\begin{equation} 
	    k^{ab}(x^{a}_{+}, x^{b}_{+}) = 0. 
	\end{equation} 
	An impulse, $j \in \mathbf{R}^l$, where $l$ is equal to the six degrees-of-freedom of an unconstrained body minus the joint's number of degrees-of-freedom, acts on both bodies to satisfy the constraint. The implicit integrator for the two-body system has the form
	\begin{equation} 
    	\begin{bmatrix} 
    	    F^{a}(x^{a}_{-}, x^{a}, x^{a}_{+}) + K^{a}(x^{a}, x^{b})^T j^{ab} \\ 
    	    F^{b}(x^{b}_{-}, x^{b}, x^{b}_{+}) + K^{b}(x^{a}, x^{b})^T j^{ab} \\ 
    	    k^{ab}(x^{a}_{+}, x^{b}_{+})
    	\end{bmatrix} = 0, \label{max_2body}
	\end{equation}
	where $K: \mathbf{X} \times \mathbf{X} \rightarrow \mathbf{R}^{l \times 6}$ is a mapping from the joint to the maximal-coordinates space and is related to the Jacobian of the joint constraint.
	
	We can generalize  \eqref{max_2body} to include additional bodies and joints. For a multi-body system with $N$ bodies and $M$ joints we define a maximal-coordinates configuration $z = (x^{(1)}, \dots, x^{(N)}) \in \mathbf{Z}$ and joint impulse $j = (j^{(1)}, \dots, j^{(M)}) \in \mathbf{J}$. We define the implicit discrete-time dynamics of the maximal-coordinates system as
	\begin{equation} 
	    F(z_{-}, z, z_{+}, j) = 0, \label{max_implicit_dynamics}
	\end{equation}
	where $F: \mathbf{Z} \times \mathbf{Z} \times \mathbf{Z} \times \mathbf{J} \rightarrow \mathbf{R}^{6N}$. In order to simulate the system we find $z_{+}$ and $j$ that satisfy \eqref{max_implicit_dynamics} for a provided $z_{-}$ and $z$ using Newton's method. 

	\begin{figure}[t]
		\centering
		\includegraphics[width=1.0\columnwidth]{figures/graph.tikz}
		\caption{Graph structure for maximal-coordinates system with 4 bodies, 3 joints, and 3 points of contact.}
		\label{graph_structure}
	\end{figure}
 
	By exploiting the mechanism's structure, we can efficiently perform root finding on \eqref{max_implicit_dynamics} (see \cite{brudigam2020linear} for additional details). This structure is manifested as a graph of the mechanism, where each body and joint is considered a node, and joints have edges connecting bodies (Fig. \ref{graph_structure}). Because the mechanism structure is known \textit{a priori}, a permutation matrix can be precomputed and used to perform efficient sparse linear algebra during simulation. For instance, in the case where the joint constraints form a system without loops, the resulting sparse system can be solved in linear time with respect to the number of links.
		
	\subsection{Variational Integrator}
	We use a specialized implicit integrator that natively handles quaternions and alleviates spurious artifacts that commonly arise from contact interactions. The dynamics are derived by approximating Hamilton's Principle of Least-Action using a simple midpoint scheme \cite{marsden2001discrete,manchester2020variational}. This approach produces \emph{variational} integrators.
 
	Each body has a linear
	\begin{equation} 
	m \frac{p_{+} - 2 p + p_{-}}{h} - h m g - A(p)^T j - h f = 0, \label{linear_integrator}
	\end{equation}
 
	and rotational
    \begin{multline}
        \sqrt{1 - \psi_+^T \psi_+} J \psi_+ + \psi_+ \times J \psi_+ \\ - \sqrt{1 - \psi^T \psi} J \psi + \psi \times J \psi - B(q)^T j - h^2 \frac{\tau}{2} = 0,
         \label{rotational_integrator}
    \end{multline}
	dynamics specified by mass $m \in \mathbf{R}_{++}$, inertia $J \in \mathbf{S}_{++}^3$, gravity $g \in \mathbf{R}^3$, and time step $h \in \mathbf{R}_{++}$. Equations (\ref{linear_integrator}, \ref{rotational_integrator}) are essentially second-order centered-finite-difference approximations of Newton's second law and Euler's equation for the rotational dynamics, respectively, where
    \begin{equation}
        q_+ = q \cdot \begin{bmatrix}
            \sqrt{1 - \psi_+^T \psi_+}~ \\
            \psi_+
        \end{bmatrix}
    \end{equation}
    is recovered from a three-parameter representation $\psi \in \mathbf{R}^3$ \cite{manchester2016quaternion}. We refer to Appendix \ref{quaternion_algebra} for quaternion conventions and algebra. Joint impulses $j \in \mathbf{J}$ have linear $A : \mathbf{R}^3 \rightarrow \mathbf{R}^{\mbox{dim}(\mathbf{J}) \times 3}$ and rotational $B : \mathbf{H} \rightarrow \mathbf{R}^{\mbox{dim}(\mathbf{J}) \times 3}$ mappings into the dynamics. The configuration of a body $x^{(i)} = (p^{(i)}, q^{(i)}) \in \mathbf{R}^3 \times \mathbf{H}$ comprises a position and orientation represented as a quaternion. Forces and torques $f, \tau \in \mathbf{R}^3$ can be applied to the bodies.

    \begin{algorithm}[t]
        \caption{Analytical Line Search For Cones}\label{cls_algo}
        \begin{algorithmic}[1]
        \Procedure{Search}{$w, \Delta, \tau^{\mbox{ort}}, \tau^{\mbox{soc}}$}
        \State $\alpha_y^{\mbox{ort}} \gets \alpha(y^{(1)}, \tau^{\mbox{ort}} \Delta^{y^{(1)}}) $ \Comment{\ref{eq:alpha_ort}}
        \State $\alpha_z^{\mbox{ort}} \gets \alpha(z^{(1)}, \tau^{\mbox{ort}} \Delta^{z^{(1)}}) $ \Comment{\ref{eq:alpha_ort}}
            
        \State $\alpha_y^{\mbox{soc}} \gets \underset{i \in \{2, \dots, n \}}{\mbox{min}} \alpha(y^{(i)}, \tau^{\mbox{soc}} \Delta^{y^{(i)}})$ \Comment{\ref{eq:alpha_soc}}
        \State $\alpha_z^{\mbox{soc}} \gets \underset{i \in \{2, \dots, n \}}{\mbox{min}} \alpha(z^{(i)}, \tau^{\mbox{soc}} \Delta^{z^{(i)}})$ \Comment{\ref{eq:alpha_soc}}
        \State \textbf{Return} $\mbox{min}(\alpha_y^{\mbox{ort}}, \alpha_z^{\mbox{ort}}, \alpha_y^{\mbox{soc}}, \alpha_z^{\mbox{soc}})$
        \EndProcedure
        \end{algorithmic}
    \end{algorithm}
    
\section{\qc{Method}} \label{dojo}

	We now introduce Dojo's contact model and custom primal-dual interior-point solver, as well as the implicit differentiation method for obtaining gradients through the solver.
    
	\subsection{Contact Dynamics Model}
	Impact and friction behaviors are modeled, along with the system's dynamics, as an NCP. This model simulates hard contact without requiring system-specific solver tuning. Additionally, contacts between a system and the environment are treated as a single graph node connected to a rigid body (Fig \ref{graph_structure}). As a result, the engine retains efficient linear-time complexity for open-chain mechanical systems.
	
	Dojo uses the rigid impact model (\ref{sdf}-\ref{impact_comp}) and in the following section we present its Coulomb friction model that utilizes an exact nonlinear friction cone.
	
	\emph{Nonlinear friction cone:}
	In contrast to the LCP approach, we utilize the optimality conditions of \eqref{mdp} in a form amenable to a primal-dual interior-point solver. The associated cone program is
	\begin{equation}
		\begin{array}{ll}
	\underset{\qc{\xi}}{\mbox{minimize}} & v^T \qc{\xi}_{(2:3)}\\
		\mbox{subject to} & \qc{\xi}_{(1)} = c_{\mathrm{f}} \gamma, \\
		& \|\qc{\xi}_{(2:3)}\|_2 \leq \qc{\xi}_{(1)},\\
		\end{array} \label{mdp_socp}
	\end{equation}
        where \qc{$\xi$ is an auxiliary vector under the nonlinear friction cone model, with elements 2 and 3 representing the friction force components}, and subscripts indicate vector indices.
	The relaxed optimality conditions for (\ref{mdp_socp}) in interior-point form are 
	\begin{align}
	    v - \eta_{(2:3)} &= 0, \\
		\qc{\xi}_{(1)} - c_{\mathrm{f}} \gamma &= 0, \\
		\qc{\xi} \circ \eta &=  \kappa \mathbf{e}, \\
		\|\qc{\xi}_{(2:3)}\|_2 \leq \qc{\xi}_{(1)}, \, \|\eta_{(2:3)}\|_2 &\leq \eta_{(1)}, 
	\end{align}
	with dual variable $\eta \in \mathbf{R}^3$ associated with the second-order-cone constraints, and central-path parameter, $\kappa \in \mathbf{R}_{+}$.
	The second-order-cone product is
	\begin{equation}
		\qc{\xi} \circ \eta = (\qc{\xi}^T \eta, \qc{\xi}_{(1)} \eta_{\qc{(2:3)}} + \eta_{(1)} \qc{\xi}_{\qc{(2:3}}), \label{soc_product}
	\end{equation}
	and
	\begin{equation} 
	    \mathbf{e} = (1, 0, \dots, 0) \label{soc_identity},
	\end{equation}
    is its corresponding identity element \cite{vandenberghe2010cvxopt}. Friction is recovered from the solution: $b = \qc{\xi}^*_{(2:3)}$. The benefits of this model are increased physical fidelity and fewer optimization variables, without substantial increase in computational cost.
    
	\emph{Nonlinear complementarity problem:}
	Systems comprising $N$ bodies and a single contact point are simulated using a time-stepping scheme that solves the feasibility problem
	\begin{align}
        \label{ncp}
    	{\mbox{find}} \quad & z_{+}, j, \gamma, \qc{\xi}, \eta \\
    	\mbox{subject to} \quad & F(z_{-}, z, z_{+}, j, \gamma, \qc{\xi}, u) = 0, \label{ncp1} \notag \\
    	& \gamma \circ \phi(z_{+}) = \kappa \textbf{1}, \notag \\
    	& \qc{\xi} \circ \eta = \kappa \mathbf{e}, \notag \\
    	& v(z, z_{+}) - \eta_{(2:3)} = 0, \notag \\
		& \qc{\xi_{(1)}} - c_{\mathrm{f}} \gamma = 0, \notag \\
    	& \gamma, \phi(z_{+}) \geq 0, \notag \\
        & \|\qc{\xi_{(2:3)}}\|_2 \leq \qc{\xi_{(1)}}, \, \|\eta_{(2:3)}\|_2 \leq \eta_{(1)}. \notag
    \end{align}
    The system's smooth dynamics $F : \mathbf{Z} \times \mathbf{Z} \times \mathbf{Z} \times \mathbf{J} \times \mathbf{R}_+ \times \mathbf{R}^2 \times \mathbf{U} \rightarrow \mathbf{R}^{6N}$ comprise linear and rotational dynamics (\ref{linear_integrator}-\ref{rotational_integrator}) for each body which are subject to inputs $u = (f^{(1)}, \tau^{(1)}, \dots, f^{(N)}, \tau^{(N)}) \in \mathbf{U}$. The contact-point tangential velocity $v : \textbf{Z} \times \textbf{Z} \rightarrow \mathbf{R}^2$ is a function of the current and next configurations (i.e., a finite-difference velocity). The central-path parameter $\kappa \in \mathbf{R}_{+}$ and target $\mathbf{e}$ \cite{vandenberghe2010cvxopt} are utilized by the interior-point solver in the following section. This formulation extends to multiple contacts.

    \begin{algorithm}[t]
        \caption{Primal-Dual Interior-Point Solver}\label{pc_algo}
        \begin{algorithmic}[1]
        \Procedure{Optimize}{$a_0, b_0, c_0, \theta, \mathcal{K}$}
        \State \textbf{Parameters}: $\tau^{\mbox{soc}}_{\mbox{max}} = 0.99, \tau_{\mbox{min}} = 0.95$
        \State $r_{\mbox{tol}} = 10^{-5}, \kappa_{\mbox{tol}} = 10^{-5}, \beta = 0.5$
        \State \textbf{Initialize}: $a = a_0, b = b_0 \in \mathcal{K}, c = c_0 \in \mathcal{K}$
        \State $r_{\mbox{vio}}, \kappa_{\mbox{vio}} \gets \Call{Violation}{w}$ \Comment{(\ref{eq:r_vio}, \ref{eq:kappa_vio})}
        
        \State \textbf{Until} $r_{\mbox{vio}} < r_{\mbox{tol}}$ \textbf{and} $\kappa_{\mbox{vio}} < \kappa_{\mbox{tol}}$ \textbf{do} 
            \State \indent \hspace{-4mm} $\Delta^{\mbox{aff}} \gets  - \bar{R}^{-1}(w; \theta) r(w; \theta, 0)$
            \State \indent \hspace{-4mm} $\alpha^{\mbox{aff}} \gets \Call{ConeSearch}{w, \Delta^{\mbox{aff}}, 1, 1}$
                
            \State \indent \hspace{-4mm} $\mu, \sigma \gets  \Call{Center}{b, c, \alpha^{\mbox{aff}}, \Delta^{\mbox{aff}}} $ \Comment{(\ref{eq:centering_0}-\ref{eq:centering_3})}
        
            \State \indent \hspace{-4mm} $\kappa \gets \mbox{max} (\sigma \mu, \kappa_{\mbox{tol}} / 5)$
            \State \indent \hspace{-4mm} $\Delta \gets  - \bar{R}^{-1}(w; \theta) r(w; \theta, \kappa)$
            \State \indent \hspace{-4mm} $\tau^{\mbox{ort}} \gets \mbox{max}(\tau_{\mbox{min}}, 1 - \mbox{max}(r_{\mbox{vio}}, \kappa_{\mbox{vio}})^2)$ 
            \State \indent \hspace{-4mm} $\tau^{\mbox{soc}} \gets \mbox{min}(\tau^{\mbox{soc}}_{\mbox{max}}, \tau^{\mbox{ort}})$ 
            \State \indent \hspace{-4mm} $\alpha \gets \Call{ConeSearch}{w, \Delta, \tau^{\mbox{ort}}, \tau^{\mbox{soc}}}$
            \State \indent \hspace{-4mm} $c^*_{\mbox{vio}}, \kappa^*_{\mbox{vio}} \gets r_{\mbox{vio}}, \kappa_{\mbox{vio}}$
            \State \indent \hspace{-4mm} $\hat{w} \gets \Call{Update}{w, \Delta, \alpha}$ \Comment{(\ref{standard_update}, \ref{quaternion_update})}
            \State \indent \hspace{-4mm} $r_{\mbox{vio}}, \kappa_{\mbox{vio}} \gets \Call{Violation}{\hat{w}}$ \Comment{(\ref{eq:r_vio}, \ref{eq:kappa_vio})}
            \State \indent \hspace{-4mm} \textbf{Until} $r_{\mbox{vio}} \leq c^*_{\mbox{vio}}$ \textbf{or} $\kappa_{\mbox{vio}} \leq \kappa^*_{\mbox{vio}}$ \textbf{do}
                \State \indent \hspace{-4mm} \indent \hspace{-4mm} $\alpha \leftarrow \beta \alpha$
                \State \indent \hspace{-4mm} \indent \hspace{-4mm} $\hat{w} \gets \Call{Update}{w, \Delta, \alpha}$ \Comment{(\ref{standard_update}, \ref{quaternion_update})}
                \State \indent \hspace{-4mm} \indent \hspace{-4mm} $r_{\mbox{vio}}, \kappa_{\mbox{vio}} \gets \Call{Violation}{\hat{w}}$ \Comment{(\ref{eq:r_vio}, \ref{eq:kappa_vio})}
            \State \indent \hspace{-4mm} \textbf{end}
            
        \State \indent \hspace{-4mm} $w \leftarrow \hat{w}$
        \State \textbf{end}
        
        \State $\partial w^* / \partial \theta \gets - \bar{R}^{-1}(w^*; \theta) \bar{D}(w^*; \theta)$ \Comment{(\ref{solution_sensitivity})}
        \State \textbf{Return} $w, \partial w^* / \partial \theta$ 
        \EndProcedure
        \end{algorithmic}
    \end{algorithm}
    
    Solving the NCP finds a maximal-coordinates state representation. In many applications it is desirable to utilize a minimial-coordinates representation (e.g., direct trajectory optimization where algorithm complexity scales with the state dimension). Dojo includes functionality to analytically convert between representations, as well as formulate and apply the appropriate chain rule in order to differentiate through a representation transformation.

    To simulate a system forward in time one step, given a control input and state comprising the previous and current configurations, solutions to a sequence of barrier problems \eqref{ncp} are found with $\kappa \rightarrow 0$. The central-path parameter has a physical interpretation as being the softness of the contact model. A value $\kappa = 0$ corresponds to exact ``hard" or inelastic contact, whereas a relaxed value produces soft contact where contact forces can occur at a distance. The primal-dual interior-point solver described in the next section adaptively decreases this parameter in order to efficiently and reliably converge to hard contact solutions. In practice, the engine is set to converge to small values \qc{(i.e. $\kappa \to 0$)} for simulation in order to simulate accurate physics. Intermediate solutions (i.e., \qc{$\kappa > 0$}) are cached and later utilized to compute smooth gradients in order to provide useful information through contact events.

    \begin{figure}[t]
		\centering
	    \includegraphics{figures/gradients.tikz}
		\caption{Gradient and \qc{dynamics} comparison between \qc{point-wise gradients} (black), randomized-smoothing gradients \cite{suh2022bundled} (orange, blue) and Dojo's analytic gradients (magenta). The dynamics for a box in the $XY$ plane that is resting on a flat surface and displaced an amount $\Delta$ by a force $f$ (top left). Randomized smoothing gradients (right column) are computed using $500$ samples with varying covariances $\Sigma$. Dojo's gradients (middle column) are computed for different values of central-path parameter $\kappa$. Compared to Dojo, the randomized smoothing method produces noisy derivatives that are many times more expensive to compute. \qc{Simulated dynamics comparisons were conducted using Dojo's results under varying values of the central-path parameter, $\kappa$. The dynamics involve a box of 1 kg mass resting on a flat surface in the $XY$ plane, displaced by a force $\Delta$ (top left). The applied force was gradually increased from 0 N to 20 N. Throughout the simulation process, considering two separate cases—one involving impact and the other involving friction—the maximum $\Delta x$ and $\Delta y$ differences between $\kappa = 1 \times 10^{-4}$ and $\kappa = 1 \times 10^{-8}$ were $1.52 \times 10^{-3}$ m and $2.56 \times 10^{-3}$ m, respectively.}}
		\label{gradient_compare}
	\end{figure}
	\subsection{Primal-Dual Interior-Point Solver} 
        \label{sec:PD-IP_solver}
	
    To efficiently and reliably satisfy \eqref{ncp}, we developed a custom primal-dual interior-point solver for NCPs with support for cone constraints and quaternions. The algorithm is largely based upon Mehrotra's predictor-corrector algorithm \cite{mehrotra1992implementation, nocedal2006numerical}, while implementing non-Euclidean optimization techniques to handle quaternions \cite{jackson2021planning} and borrowing features from CVXOPT \cite{vandenberghe2010cvxopt} to handle cones. 
	
    The primary advantages of this algorithm are the correction to the classic Newton step, which can greatly reduce the iterations required by the solver (often halving the total number of iterations), and feedback on the problem's central-path parameter that helps avoid premature ill-conditioning and adaptively drives the complementarity violation to zero in order to reliably simulate hard contact.

    \qc{The solver functions as a systematic procedure that iteratively refines the solution while maintaining feasibility within the prescribed cone constraints. The primal-dual framework provides a structured pathway to the solution, and the interplay between the affine (predictor) and corrector steps ensures steady progress towards eliminating both constraint and complementarity violations. The analytical line search offers a principled means of selecting step sizes that keep the iterates strictly within the cone, while the specialized handling of non-Euclidean variables ensures stable and accurate updates. By continuously monitoring violations and adjusting parameters accordingly, the approach reliably converges to a solution that meets predefined tolerances. As a result, this solver is capable of addressing a broad range of problems—from those with simple linear conditions to complex, nonlinear scenarios—while providing accurate solutions and informative gradients.}
	
    \subsubsection{\qc{Problem formulation}}
    The solver aims to satisfy instantiations of the following problem
    \begin{equation}
    	\begin{array}{ll}
    	\underset{}{\mbox{find}} & a, b, c \\
    	\mbox{subject to} & E(a, b, c; \theta) = 0, \\
    	& b \circ c = \kappa \mathbf{e}, \\
    	& b, c \in \mathcal{K},
    	\end{array} \label{ncp_abstract}
    \end{equation}
    with decision variables $a \in \mathbf{R}^{n_{a}}$ and $b, c \in \mathbf{R}^{n_{\mathcal{K}}}$, equality-constraint set $E : \mathbf{R}^{n_{a}} \times \mathbf{R}^{n_{\mathcal{K}}} \times \mathbf{R}^{n_{\mathcal{K}}} \times \mathbf{R}^{n_{\theta}} \rightarrow \mathbf{R}^{n_a + n_{\mathcal{K}}}$, problem data $\theta \in \mathbf{R}^{n_{\theta}}$; and where $\mathcal{K}$ is the Cartesian product of positive-orthant and second-order cones \cite{boyd2004convex}. 
	
	Interior-point methods aim to satisfy a sequence of relaxed problems with $\kappa > 0$ and $\kappa \rightarrow 0$ in order to reliably converge to a solution of the original problem (i.e., $\kappa = 0$). This continuation approach, \qc{though it makes our interior point solver hard to warm-start,} helps avoid premature ill-conditioning and is the basis for numerous convex and nonconvex interior-point solvers \cite{nocedal2006numerical}.

    \begin{table}
		\centering
		\caption{Contact violation for Atlas drop (Fig.~\ref{atlas_drop}). Comparison between Dojo and MuJoCo for foot contact penetration  (millimeters) with the floor for different time steps (seconds). Dojo strictly enforces no penetration. When Atlas lands, its feet remains above the ground by an infinitesimal amount. In contrast, MuJoCo exhibits significant penetration through the floor (i.e., negative values).}
		\begin{tabular}{c c c c}
			\toprule
			    \textbf{Time Step} & \textbf{0.1} & \textbf{0.01} & \textbf{0.001} \\
			\toprule
			MuJoCo & \mbox{failure} & \textminus 28 & \textminus 46 \\
			Dojo & \textbf{+1e{-}12} & \textbf{+1e{-}7} & \textbf{+8e{-}6} \\
			\toprule
		\end{tabular}
		\label{contact_violation_results}
	\end{table}
 
    The LCP formulation is a special-case instantiation of \eqref{ncp_abstract} where the constraint set is affine in the decision variables and the cone is the positive orthant. Most general-purpose solvers for LCP problems rely on active-set methods that strictly enforce $\kappa = 0$ at each iteration. Consequently, these solvers generate non-informative gradient information (see Section \ref{sec:gradients}).  In contrast our interior point solver can give informative gradients with a smoothing effect related to the size of $\kappa$.
    \subsubsection{\qc{Residual and Jacobians}}
    The interior-point solver aims to find a fixed point for the residual
    \begin{align}
        r(w; \theta, \kappa) &= \begin{bmatrix}
            E(w;\theta) \\
            b^{(1)} \circ c^{(1)} - \kappa \mathbf{1}\\
            \vdots \\
            b^{(n_{\mathcal{K}})} \circ c^{(n_{\mathcal{K}})} - \kappa \mathbf{e}\\
        \end{bmatrix}, \label{residual}
    \end{align}
    while respecting the cone constraints.
    The Jacobian of this residual with respect to the decision variables
    \begin{equation}
            R(w; \theta) = \frac{\partial r(w; \theta, \cdot)}{\partial w}, \label{var_jacobian}
    \end{equation}
    is used to compute a search direction. For convenience, we denote $w = (a, b, c)$. After a solution $w^*(\theta, \kappa)$ is found, the Jacobian of the residual with respect to the problem data
    \begin{equation}
            D(w; \theta) = \frac{\partial r(w; \theta, \cdot)}{\partial \theta} , \label{data_jacobian}
    \end{equation}
    is used to compute the sensitivity of the solution. These Jacobians are not explicitly dependent on the central-path parameter.
 
    The non-Euclidean properties of quaternion variables are handled with modifications to these Jacobians \eqref{var_jacobian} and \eqref{data_jacobian} by right multiplying each with a matrix $H$ containing attitude Jacobians \cite{jackson2021planning} corresponding to the quaternions in $x$ and $\theta$, respectively 
    \begin{align}
        \bar{R}(w; \theta) &= R(w; \theta) H_{R}(w), \\ 
        \bar{D}(w; \theta) &= D(w; \theta) H_{D}(\theta).
    \end{align}
    Euclidean variables have corresponding identity blocks. This modification accounts for the implicit unit-norm constraint on each quaternion variable and improves the convergence behaviour of the solver.
 
    \subsubsection{\qc{Cones}}
	The generalized inequality, cone-product operator, and target for the $n$-dimensional positive orthant are
    \begin{align}
        \mathbf{R}^n_{++} &= \{a \in \mathbf{R}^n \, | \, a_{(i)} > 0, \, i = 1, \dots, n\}, \\
        a \circ b &= (a_{(1)} b_{(1)}, \dots, a_{(n)} b_{(n)}), \\
        \mathbf{e} &= \mathbf{1}.
    \end{align}
    For the second-order cone they are 
    \begin{align}
    	\mathcal{Q}^n &= \{(a_{(1)}, a_{(2:n)}) \in \mathbf{R} \times \mathbf{R}^{n-1}\, | \, \|a_{(2:n)}\|_2 \leq a_{(1)} \}, \\
    	a \circ b &= (a^T b, a_{(1)} b_{(2:n)} + a_{(1)} b_{(2:n)}),\\ \mathbf{e} &= (1, 0, \dots, 0).
	\end{align}
	The solver utilizes the Cartesian product
	\begin{equation} 
	    \mathcal{K} = \mathbf{R}^n_{+} \times \mathcal{Q}^{l_1}_1 \times \dots \times \mathcal{Q}^{l_j}_j,
	\end{equation}
	of the $n$-dimensional positive orthant and $j$ second-order cones, each of dimension $l_i$.
 
    \subsubsection{\qc{Analytical line search for cones}}
    To ensure the cone variables strictly satisfy their constraints, a cone line search is performed for a candidate search direction. For the update
    \begin{align}
        y \leftarrow y + \alpha \Delta,
    \end{align}
    with step size $\alpha$ and search direction $\Delta$, the solver finds the largest $\alpha \in [0, 1]$ such that $y + \alpha \Delta \in \mathcal{K}$. The step-size is computed analytically for the positive orthant
    \begin{align}
        \alpha = \mbox{min} \left(1,  \underset{k | \Delta_{(k)} < 0}{\mbox{max}} \Big \{ -\frac{y_{(k)}}{\Delta_{(k)}} \Big \} \right), 
        \label{eq:alpha_ort}
    \end{align}
    and second-order cone
    \begin{align}
        \nu &= y_{(1)}^2 - y_{(2:k)}^T y_{(2:k)}, \\
        \zeta &= y_{(1)} \Delta_{(1)} - y_{(2:k)}^T \Delta_{(2:k)}, \\
        \rho_{(1)} &= \frac{\zeta}{\nu}, \\
        \rho_{(2:k)} &= \frac{\Delta_{(2:k)}}{\sqrt{\nu}} - \frac{\frac{\zeta}{\sqrt{\nu}} + \Delta_{(1)}}{y_{(1)}\sqrt{\nu} + \nu} y_{(2:k)}, \\
        \alpha &= \begin{cases}
            \mbox{min} \left( 1, \frac{1}{\|\rho_{(2:k)}\|_2 - \rho_{(1)}} \right), \quad \|\rho_{(2:k)}\|_2 > \rho_{(1)}, \\
            1, \quad \mbox{otherwise}.
            \end{cases}
        \label{eq:alpha_soc}
    \end{align}
    The line search over all individual cones is summarized in Algorithm \ref{cls_algo}.

    \begin{figure}[t]
		\begin{center}
            \begin{tikzpicture}
                \draw (0, 0) node[inner sep=0] {\includegraphics[width=\linewidth]{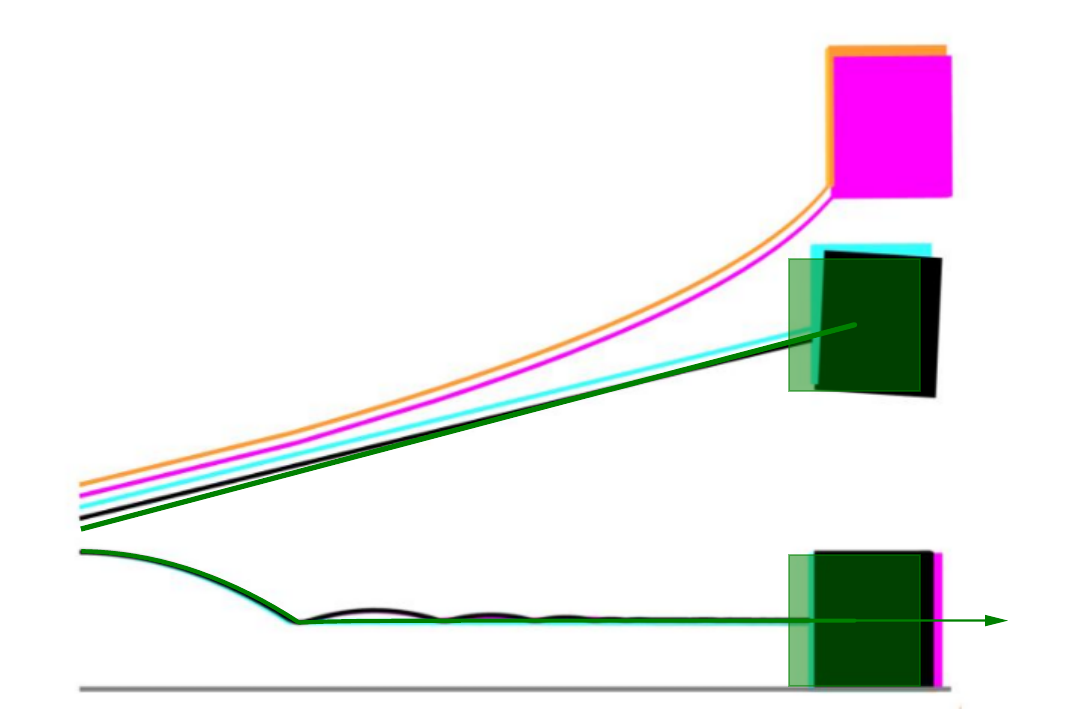}};
                \draw (+0.0, +1.7) node {top view};
                \draw (+0.0, -1.4) node {side view};
            \end{tikzpicture}
		\end{center}
        \caption{Velocity drift resulting from friction-cone approximation. Comparison between a box sliding with approximate cones having four vertices implemented in MuJoCo (magenta) and Dojo (orange) versus MuJoCo's (black) and Dojo's (blue) \qc{and PyBullet's (green)} nonlinear friction cones. Dojo's nonlinear friction cone gives the physically correct straight line motion, while linear friction-cone approximations lead to lateral drift. MuJoCo's nonlinear friction cone exhibits a minor rotational drift. \qc{PyBullet's nonlinear friction cone delivers correct straight line motion, but exerts greater friction force on the box leading to a shorter trajectory than Dojo and MuJoCo.}}
        \label{velocity_drift}
    \end{figure}
    
    \subsubsection{\qc{Candidate update}} 
    The variables are partitioned: $a = (a^{(1)}, \dots, a^{(p)})$, where $i = 1$ are Euclidean variables and $i = 2, \dots, p$ are each quaternion variables; and $b = (b^{(1)}, \dots, b^{(n)})$, $c = (c^{(1)}, \dots, c^{(n)})$, where $j = 1$ is the positive-orthant and the remaining $j = 2, \dots, n$ are second-order cones.
    For a given search direction, updates for Euclidean and quaternion variables are performed. The Euclidean variables in $a$ use a standard update 
    \begin{equation} 
        a^{(1)} \leftarrow a^{(1)} + \alpha \Delta^{(1)}. \label{standard_update}
    \end{equation}
    For each quaternion variable, the search direction exists in the space tangent to the unit-quaternion hypersphere and is 3-dimensional. The corresponding update for $i = 2, \dots, p$ is 
    \begin{equation}
        a^{(i)} \leftarrow L(a^{(i)}) \varphi(\alpha \Delta^{(i)}), \label{quaternion_update}
    \end{equation}
    where $L : \mathbf{H} \rightarrow \mathbf{R}^{4 \times 4}$ is a matrix representing a left-quaternion matrix multiplication, and $\varphi : \mathbf{R}^3 \rightarrow \mathbf{H}$ is a mapping to a unit quaternion. The standard update \eqref{standard_update} is used for the remaining decision variables $b$ and $c$.
 
    \subsubsection{\qc{Violation metrics}}
    Two metrics are used to measure progress: 
    (i) the constraint violation
    \begin{align}
        r_{\mbox{vio}} = \| r(w; \theta) \|_{\infty},
        \label{eq:r_vio}
    \end{align}
    and (ii) complementarity violation
    \begin{align}
        \kappa_{\mbox{vio}} = \underset{i}{\mbox{max}} \{\| b^{(i)} \circ c^{(i)} \|_{\infty}\}.
        \label{eq:kappa_vio}
    \end{align}
    The problem \eqref{ncp_abstract} is considered solved when $r_{\mbox{vio}} < r_{\mbox{tol}}$ and $\kappa_{\mbox{vio}} < \kappa_{\mbox{tol}}$.
	
    \subsubsection{\qc{Centering}}
    The solver adaptively relaxes \eqref{ncp_abstract} by computing the centering parameters $\mu$ and $\sigma$. These values provide an estimate of the cone-constraint violation and determine the value of the central-path parameter that a correction step will aim to satisfy. These values rely on the degree of the cone \cite{vandenberghe2010cvxopt},
    \begin{equation}
        \mbox{\textbf{deg}}(\mathcal{K}) = \sum_{i=1}^{n_{\mathcal{K}}} \mbox{\textbf{deg}}(\mathcal{K}^{(i)}) = \mbox{\textbf{dim}}(\mathcal{K}^{(1)}) + n_{\mathcal{K}} - 1 \label{eq:centering_0},
    \end{equation}
    the complementarity violations,
    \begin{equation}
        \mu = \frac{1}{\mbox{\textbf{deg}}(\mathcal{K})} \sum_{i = 1}^{n_{\mathcal{K}}} (b^{(i)})^T c^{(i)}, \label{eq:centering_1}
    \end{equation}
    and affine complementarity violations, 
    \begin{equation}
        \mu^{\mbox{aff}} = \frac{1}{\mbox{\textbf{deg}}(\mathcal{K})} \sum_{i = 1}^{n_{\mathcal{K}}} (b^{(i)} + \alpha \Delta^{b^{(i)}})^T (c^{(i)} + \alpha \Delta^{c^{(i)}}), \label{eq:centering_2}
    \end{equation}
     as well as their ratio,
    \begin{equation}
        \sigma = \mbox{min}\left(1, \mbox{max} \left(0, \frac{\mu^{\mbox{aff}}}{\mu} \right) \right)^3.
        \label{eq:centering_3}
    \end{equation}
    As the algorithm makes progress, it aims to reduce these violations.
 
    \subsubsection{\qc{Algorithm}}
    The interior-point algorithm used to solve \eqref{ncp_abstract} is summarized in Algorithm \ref{pc_algo}. Additional tolerances $\tau \in [0.9, 1]$ are used to improve numerical reliability of the solver. The algorithm parameters include $\tau^{\mbox{soc}}_{\mbox{max}}$ to prevent the iterates from reaching the boundaries of the cones too rapidly during the solve, $\tau_{\mbox{min}}$ to ensure we are aiming at sufficiently large steps, and $\beta$ is the decay rate of the step size $\alpha$ during the line search. In practice, $r_{\mbox{tol}}$ and $\kappa_{\mbox{tol}}$ are the only parameters the user might want to tune.
    
    Finally, the algorithm outputs a solution $w^*(\theta, \kappa)$ that satisfies the solver tolerance levels and, optionally, the implicit gradients of the solution with respect to the problem parameters $\theta$.
  
    For an instance of problem \eqref{ncp_abstract}, the algorithm is provided problem data and an initial point, which is projected to ensure that the cone variables are initially feasible with some margin. Next, an affine search direction (i.e., predictor) is computed that aims for zero complementarity violation. Using this direction, a cone line search is performed followed by a centering step that computes a target relaxation for the computation of the corrector search direction. A second cone line search is then performed for this new search direction. A subsequent line search is performed until either the constraint or complementarity violation is reduced. The current point is then updated, a new affine search direction is computed, and the procedure repeats until the violations satisfy the solver tolerances.

    \subsection{Gradients}
    \label{sec:gradients}
	\qc{Dojo simulates a user-tuneable smoothed approximation of hard contact dynamics. This approach allows us to compute gradients that are more informative in the presence of contacts by enabling a force-from-a-distance mechanism, as discussed in references \cite{posa2014direct} and \cite{suh2022bundled}. } As previously discussed, interior-point methods optimize a sequence of smooth barrier sub-problems, where the degree of smoothing is parameterized by the central-path parameter $\kappa$. \mac{Differentiating at a large value of $\kappa$ gives more contact smoothing, with more informative gradients but less accurate solutions, while differentiating at small $\kappa$ values gives less smoothing, less informative gradients, but better physical fidelity.  The chosen intermediate solution, $w^*(\theta, \kappa > 0)$, is} differentiated using the implicit function theorem \eqref{solution_sensitivity} to compute smooth \textit{implicit gradients}. In practice, we find that these gradients greatly improve the performance of gradient-based optimization methods, consistent with the long history of interior-point methods. Dojo's gradients are compared with \qc{point-wise gradients} and randomized smoothing in Fig. \ref{gradient_compare}. \qc{Since $\kappa$ represents a tradeoff between simulation accuracy and gradient smoothness, we evaluate the effect of $\kappa$ on simulation accuracy.  We compare the simulation results of different $\kappa$ values in Fig. \ref{gradient_compare}.} 
    
    The problem data for each simulation step includes: the previous and current configurations, control input, and additional terms like the time step, friction coefficients, and parameters of each body. 
 
    \subsection{Implementation}
    An open-source implementation, \texttt{Dojo.jl}, written in Julia, is available and a Python interface, \texttt{dojopy}, is also included. These tools, and the experiments, are available at, 
    \begin{center}
    \url{https://www.github.com/dojo-sim/Dojo.jl}.
    \end{center}
    
	\section{Results} \label{examples}
	Dojo's capabilites are highlighted through a collection of examples, including: simulating physical phenomena, gradient-based planning with trajectory optimization, policy optimization, system identification, \mac{and sim-to-real gap evaluation with robot hardware}. The current implementation supports point, sphere, and capsule collisions with flat surfaces with a pre-existing collision detection module (which is outside the scope of this work). All of the experiments were performed on a computer with an Intel Core i9-10885H processor and 32GB of memory.
 
	\subsection{Simulation}

    \begin{figure}[t]
    \centering
        \includegraphics[width=0.25\textwidth]{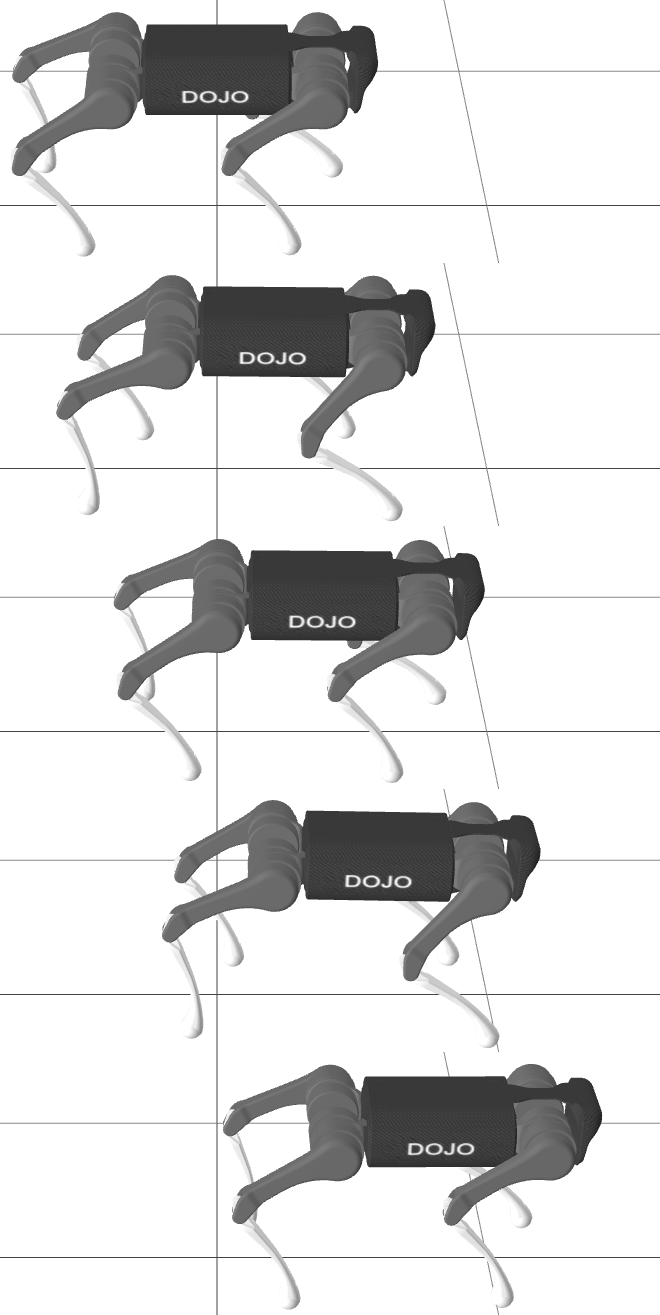}
		\caption{Locomotion plan for quadruped generated using trajectory optimization. Time progresses top to bottom.}
		\label{trajopt_vis}
	\end{figure}
 
	The simulation accuracy of Dojo and MuJoCo is compared in a number of illustrative  scenarios.
    
	\emph{Impact constraints comparison:} 
	The Atlas humanoid is simulated dropping onto a flat surface (Fig. \ref{atlas_drop}). The system comprises 31 bodies, resulting in 403 maximal-coordinates states, and has 36 actuated degrees-of-freedom. Each foot has four contact points. A comparison with MuJoCo is performed measuring penetration violations with the floor for different simulation rates (Table \ref{contact_violation_results}). The current implementation of Dojo simulates this system in real time at 65 Hz.
    
	\emph{Friction-cone comparison:}
    The effect of friction-cone approximation is demonstrated by simulating a box that is initialized with lateral velocity before impacting and sliding along a flat surface. The complementarity problem with $P$ contact points requires $2 P (1 + 2 d)$ decision variables for contact and a corresponding number of constraints, where $d$ is the degree of parameterization (e.g., double parameterization: $d=2$). For a pyramidal approximation, in the probable scenario where its vertices are not aligned with the direction of motion, velocity drift occurs for a linearized cone implemented in Dojo and MuJoCo. \qc{Meanwhile, though MuJoCo and PyBullet also can use nonlinear friction cones, compared to Dojo, MuJoCo’s nonlinear friction cone exhibits a minor rotational drift and PyBullet’s nonlinear friction cone exerts greater friction force on box that leads to a shorter trajectory} (Fig. \ref{velocity_drift}). While it is possible to reduce such artifacts by increasing the number of vertices in the approximation of the second-order cone, this increases the computational complexity. Such approximation is unnecessary in Dojo as we handle the exact nonlinear cone constraint efficiently and reliably with optimization tools from cone programming; the result is accurate sliding.
 
    \begin{table}[t]
        \centering
		\caption{Planning results. Comparison of final cost value, goal constraint violation, and total number of iterations for a collection of systems optimized with iterative LQR \cite{li2004iterative} using Dojo (D) with implicit gradients or MuJoCo (M) with finite-difference gradients.}
		\begin{tabular}{c c c c}
			\toprule
			\textbf{System} & \textbf{Cost} & \textbf{Violation} & \textbf{Iterations}\\
			\toprule
			box right (D) & 14.5 & 3e{-}3 & \textbf{30} \\
			box right (M) & \textbf{13.5} & \textbf{3e{-}3} & 95 \\
			\hline
			box up (D) & \textbf{14.5} & \textbf{3e{-}3} & \textbf{106} \\
			box up (M) & \mbox{failure} & 1.0 & -\\
			\hline
			hopper (D) & \textbf{8.9} & \textbf{1e{-}3} & 96 \\
			hopper (M) & 26.7 & 2e{-}3 & \textbf{66} \\
			\hline
			quadruped (D) & 2e{-}2 & 3e{-}4 & 20 \\
			\toprule
		\end{tabular}
		\label{trajopt_results}
    \end{table}
 
\begin{figure}[t]
		\begin{center}
	        \includegraphics[width=0.8\columnwidth]{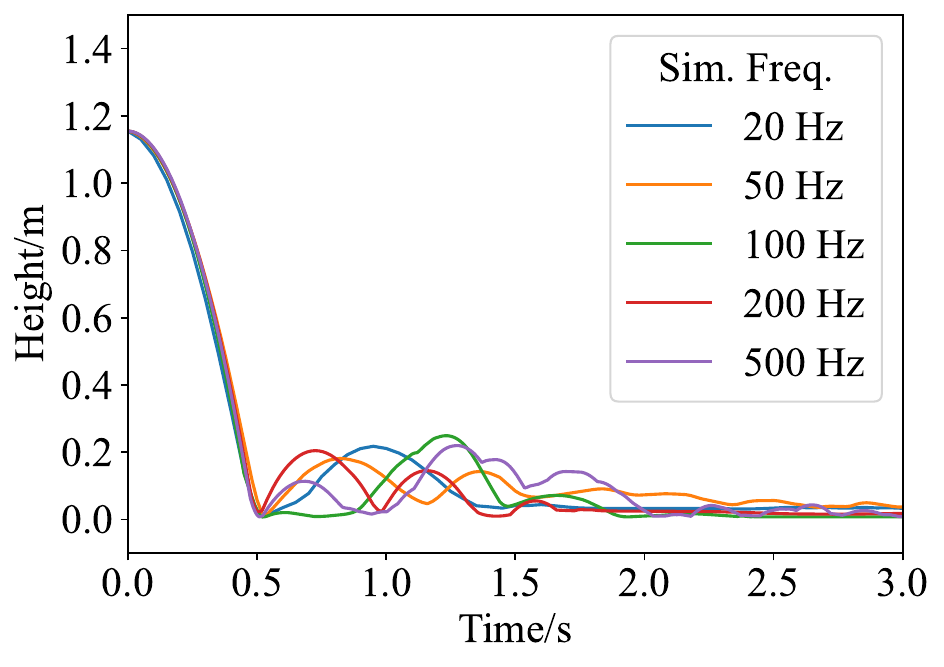}
			\includegraphics[width=0.8\columnwidth]{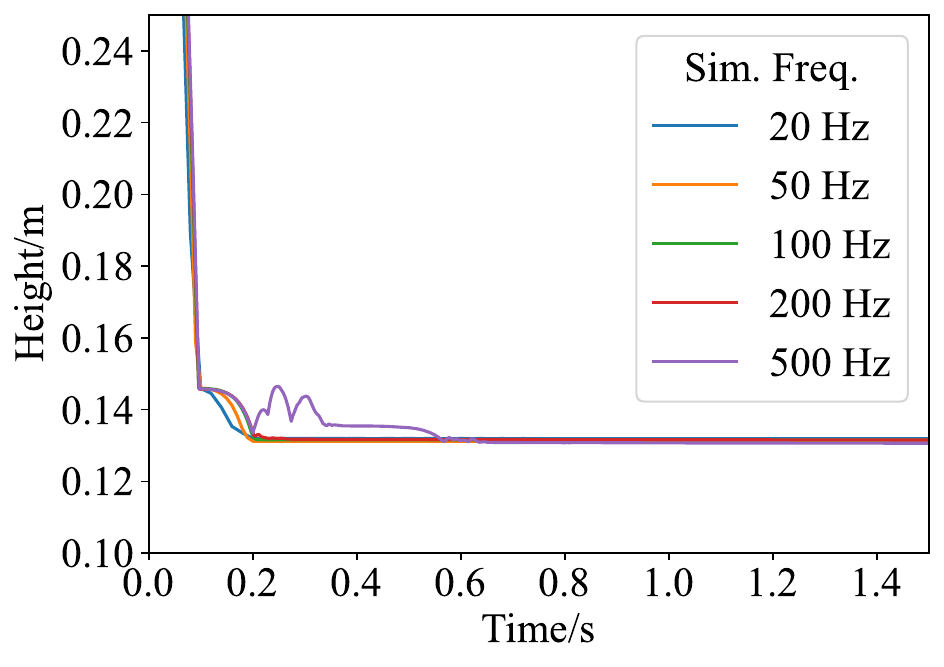}
		\end{center}
		\caption{\qc{Torso height over time under different simulation frequency for Atlas (top) drop and A1 (bottom) drop experiments.}}
		\label{time_step}
    \end{figure}

    \qc{
    \emph{Simulation Stability at Low Frequencies} To validate the stability of the primal-dual interior point solver under different simulation frequencies, we conduct Atlas drop (Fig.~\ref{atlas_drop}) and similar quadrupedal drop experiments under different simulation frequencies. During the simulations, we recorded the robots' torso heights over time. The simulation results can be found in Fig.~\ref{time_step}. Under a large span of simulation frequency from 20-500 Hz, both Atlas drop and quadrupedal drop deliver similar simulation results with small reasonable deviations. The test results demonstrate that Dojo preserves simulation fidelity at low frequency, even through contact events.} 

    \qc{\emph{Computation time:} One of the primary challenges with differentiable simulators is their computation speed. Table \ref{table:time_benchmark} presents a benchmark comparing the computation time for forward simulation and gradient calculation across four simulators on four different robot types. Each test involved simulating 1000 steps with a time step of 0.01s, using randomly generated actions. MuJoCo's gradients are calculated with built-in finite differentiation function, Drake utilizes its randomized smoothing gradients calculation function, while Brax calculates gradients through auto-differentiation. 
    
    Among all four simulators, MuJoCo exhibited a clear advantage in computation speed. However, Dojo significantly outperforms another differentiable simulator, Brax, and delivers a comparable performance as Drake. This advantage stems from (i) the maximal-coordinate dynamics model, which is highly effective in handling complex systems with multiple links and joints, as discussed in Section \ref{maximal-coordinates}, and (ii) Dojo's implicit differentiation method, which avoids the $\textit{O}(n^2)$ complexity of finite difference techniques in the action and observation spaces. }

\begin{figure}[t]
        \begin{center}
            \includegraphics[width=1.0\columnwidth]{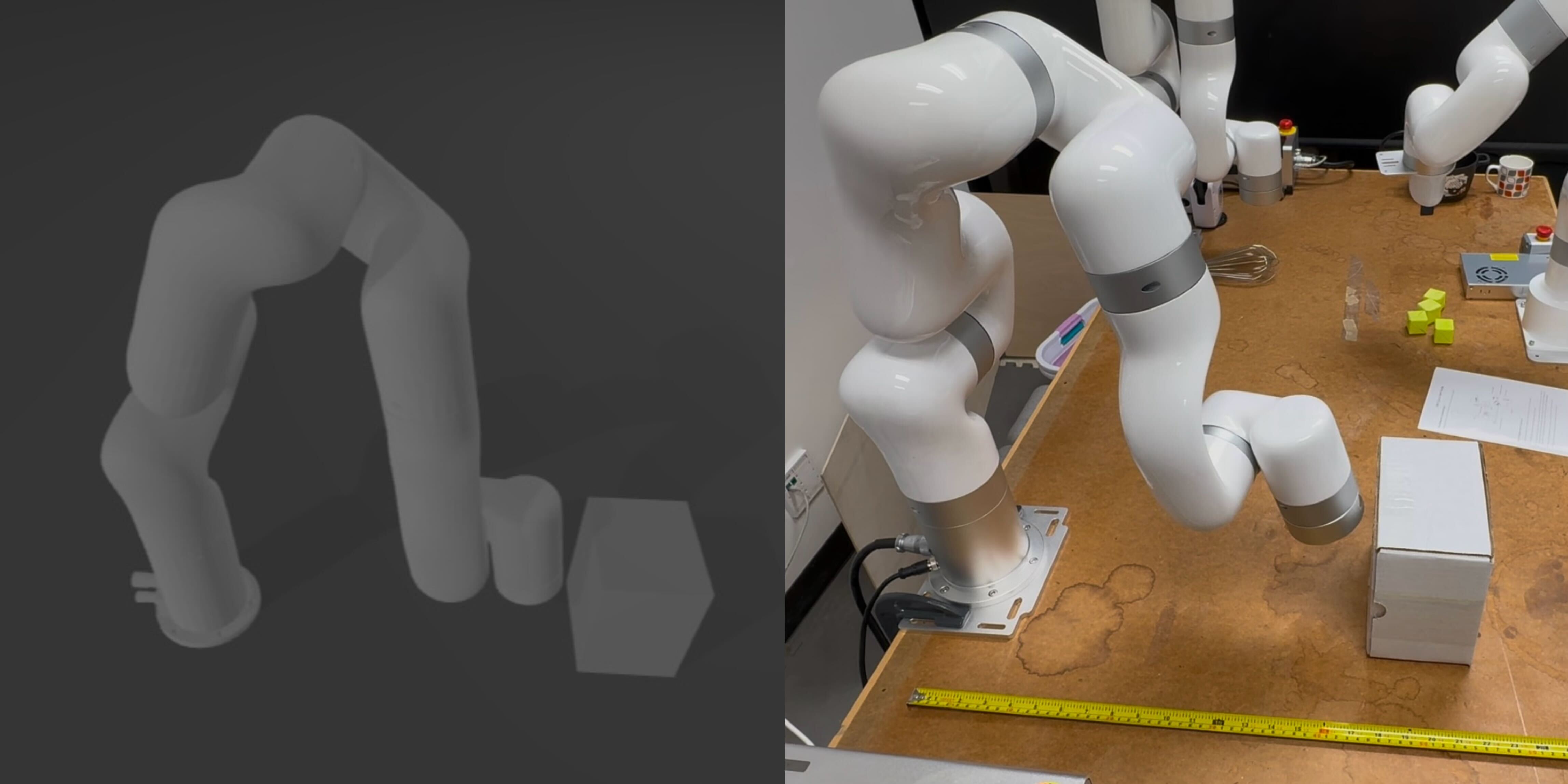}
        \end{center}
        \caption{\qc{Simulation and real robot comparison under physical contacts and frictions. Robot arm pushing box experiment scenarios.}}
        \label{figure:sim-to-real}
    \end{figure}

    \qc{\emph{Sim-to-Real gap evaluation:} To assess the sim-to-real transfer capabilities and fidelity of Dojo, we conducted a series of box pushing experiments using a 6-axis xArm manipulator and a 0.5 kg rectangular box (11 cm by 13 cm by 21 cm) positioned at initial x-axis distances of 30, 35, 40, 45, and 50 cm from the manipulator's base, the simulation and real experimental scenario can be seen in Fig. \ref{figure:sim-to-real}. In both the simulated and real-world trials, identical proportional-derivative (PD) gains and joint-space commands were applied. We record the box's final location and flipping status of each experiment (real world) or simulation (Dojo), the results can be found in Table \ref{table:sim-to-real}. After the pushing action, the positional discrepancy in the box’s final location between simulation and reality averaged approximately 0.5 cm (1.25\%). Apart from positional accuracy, we also examined the flipping behavior of the box, which is sensitive to the precise point of contact and influenced by complex frictional and contact forces. Dojo’s predictions of flipping outcomes closely matched those observed in the physical experiments, indicating that the simulator can capture the subtle and intricate dynamics involved in frictional contact scenarios. Taken together, these results give evidence of Dojo's ability to reproduce physical phenomena, thereby supporting robust sim-to-real transfer in robotic manipulation tasks.}

\begin{table}[]
\centering
\caption{\qc{Robot arm pushing box experiment results for Sim-to-Real gap evaluation. Box was placed at different locations on the table with different initial x-direction distances to robot arm's base. Both simulation and real robot are controlled by the same PD controller to track the same joint space at 100 Hz.}}
\qc{
\begin{tabular}{ccccccc}
\toprule
\multicolumn{2}{c}{\textbf{Initial Distance (cm)}}         & \textbf{30} & \textbf{35} & \textbf{40} & \textbf{45} & \textbf{50} \\ \midrule
\multirow{2}{*}{\textbf{Real}}    & Box Flip               & Yes         & Yes         & Yes         & Yes         & No          \\
                                  & Final Distance (cm)    & 46.9        & 52.7        & 55.1        & 61.2        & 50          \\ \midrule
\multirow{2}{*}{\textbf{Sim}}     & Box Flip               & Yes         & Yes         & Yes         & Yes         & No          \\
                                  & Final Distance (cm)    & 45.7        & 52.3        & 55.2        & 62          & 50          \\ \midrule
\multicolumn{2}{c}{\textbf{Sim-to-Real Distance Gap (cm)}} & 1.2         & 0.4         & 0.1         & 0.8         & 0\\          
\toprule
\end{tabular}
}
\label{table:sim-to-real}
\end{table}

    \renewcommand{\arraystretch}{1} 
      \captionsetup{
    	skip=5pt, position = bottom}
    \begin{table*}
        \small
        \caption{\qc{Computation time benchmark results. Comparison of computation time of forward simulation plus gradient calculation for different simulators on different types of robots. Simulation time step is 0.01 s, each test was simulated for 1000 steps, with randomly generated actions. MuJoCo's gradients are calculated with built-in finite differentiation function, Drake utilizes its randomized smoothing gradients calculation function, while Brax calculates gradients through auto-differentiation.}}
        \label{table:time_benchmark}
        \centering
        \qc{
        \begin{tabular}{ccccc}
        \toprule
        \multirow{2}{*}{\textbf{Simulator}} & \multicolumn{4}{c}{\textbf{Simulation Time of Different Robots [s]}}                                                                   \\
                                            & Humanoid         & Unitree A1       & Franka Panda    & Skydio X2         \\ \midrule
        MuJoCo                              & 1.512 $\pm$ 0.045         & 1.114 $\pm$ 0.007         & 0.335 $\pm$ 0.001          & 0.047 $\pm$ 0.002          \\
        Drake                               & 5.463 $\pm$ 0.078          & 3.870 $\pm$ 0.024         & 2.352 $\pm$ 0.055          & 0.571 $\pm$ 0.011         \\
        Brax                                & 6.975 $\pm$ 0.485           & 11.064 $\pm$ 0.521         & 10.954 $\pm$ 0.395        & 7.953 $\pm$ 0.484           \\
        Dojo                                & 1.750 $\pm$ 0.135 & 5.235 $\pm$ 0.071         & 1.159 $\pm$ 0.077 & 0.807 $\pm$ 0.003   \\ \bottomrule        
        \end{tabular}}
    \end{table*}

\begin{figure}[t]
        \begin{center}
            \includegraphics[width=1.0\columnwidth]{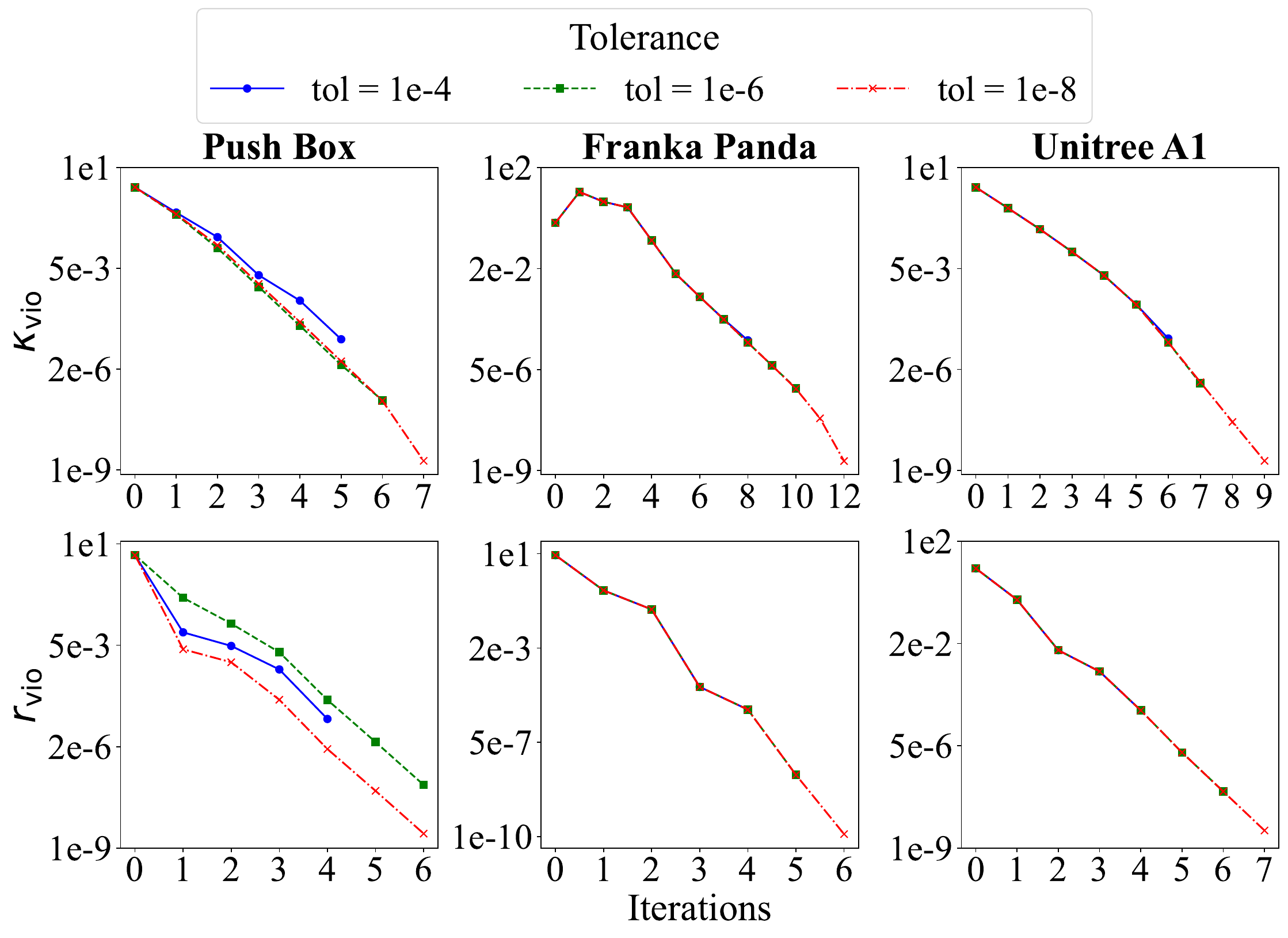}
        \end{center}
        \caption{\qc{Plots of residual values $\kappa_{vio}$ and $r_{vio}$  versus iteration under different tolerance for three different robots.}}
        \label{residual_plot}
    \end{figure}

\begin{figure}[t]
        \begin{center}
            \includegraphics[width=1.0\columnwidth]{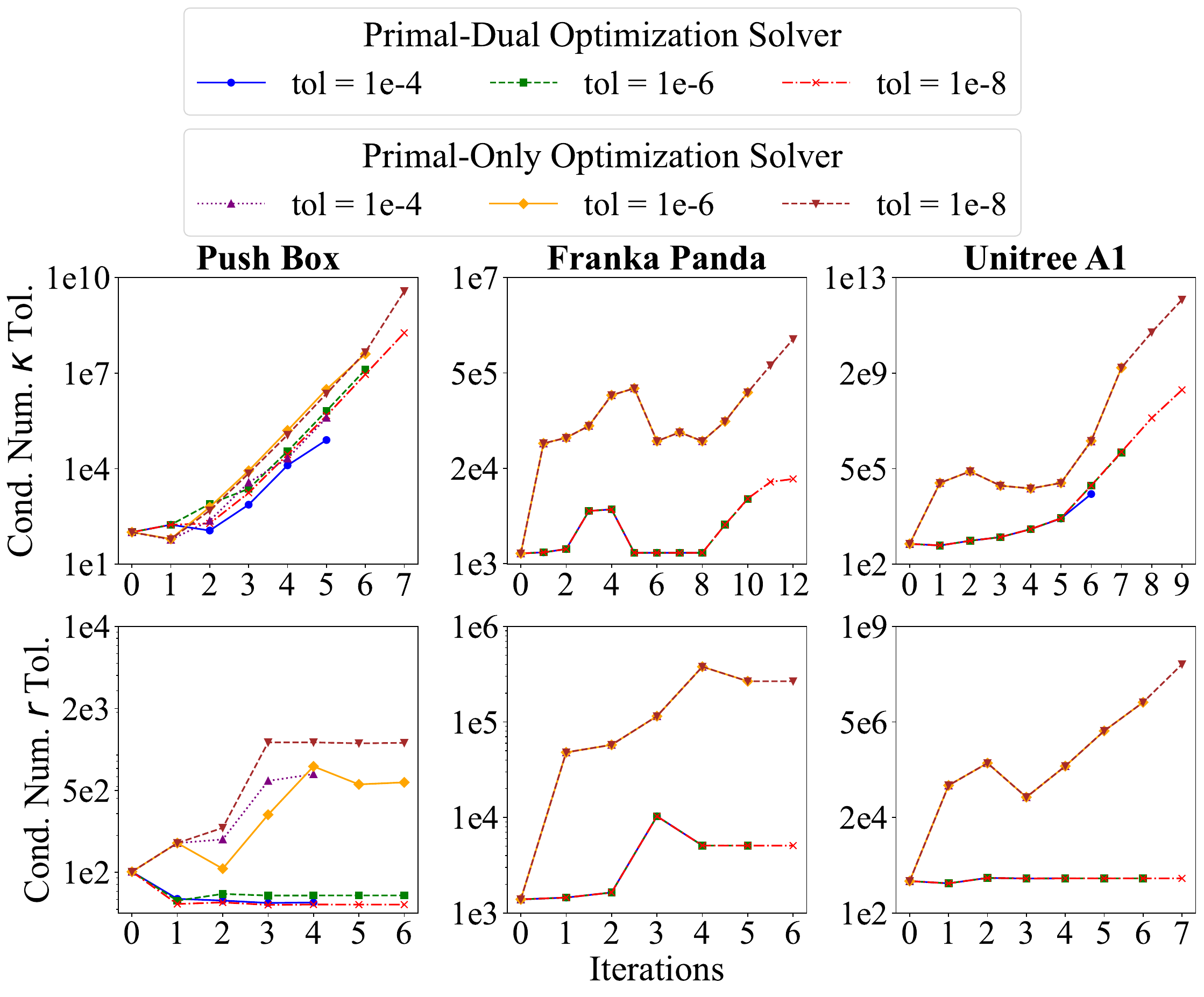}
        \end{center}
        \caption{\qc{Plots of condition numbers versus iteration under different $\kappa$ and $r$ tolerance for three different robots.}}
        \label{condition_plot}
    \end{figure}

    \qc{\emph{Convergence study: }Dojo's solver reduces the constraint violation $r_{vio}$ and complementarity violation $\kappa_{vio}$ until both residual values are smaller than prescribed tolerances. As the problem is nonconvex, it is important to analyze Dojo's convergence performance across different robots under different conditions. We simulate three robots in Dojo and record their $\kappa_{vio}$ and $r_{vio}$ as well as the solver's condition numbers over iterations under different tolerance settings. Meanwhile, to further substantiate Dojo's ability to avoid ill-conditioning, we also compare Dojo with a primal-only interior point solver \cite{castro2022unconstrained} on condition numbers over iterations under different tolerance settings. The convergence study results are shown in Fig. \ref{residual_plot} and \ref{condition_plot}. 
    
    When testing with different $\kappa_{tol}$, we fix $r_{tol}$ at $1\times10^{-8}$, while as $\kappa_{vio}$ is the main bottleneck of the solver, we relax $\kappa_{tol}$ to 0.1 when testing with different $r_{tol}$.  We do the same for both residual values and condition number experiments. The experiment results demonstrate that Dojo's primal-dual interior-point solver can converge within 15 iterations for all three robots. The residual decreases reliably for different tolerance settings, showing that Dojo's solver has strong numerical stability. For condition number experiments, the general trend is that condition number increases with the iteration for both methods, as expected. Specifically, for Dojo, under relaxed $\kappa_{tol}$ (0.1), the condition number stays low (smaller than $1\times10^{-4}$) with iterations for all three robots. Meanwhile, under rigorous $\kappa_{tol}$ ($1\times10^{-8}$), the condition number was higher than that under relaxed $\kappa_{tol}$, but still lies at a reasonable level for a nonconvex optimization problem (smaller than $1\times10^{9}$). Moreover, the primal-only solver's condition number is much higher than the proposed method. Under different tolerance settings, at the last iteration of each experiment, the primal method's condition number is 2 to $1\times10^{4}$ times of Dojo's condition number, showing that Dojo's primal-dual interior-point solver has an advantage in avoiding numerical ill-conditioning.}

	\subsection{Planning} 
	
	\qc{We utilize iterative LQR by providing implicit gradients from Dojo} \cite{howell2022trajectory} to perform trajectory optimization on three systems: planar box, hopper, and quadruped. A comparison is performed with MuJoCo and finite-difference gradients. The results are visualized for the quadruped in Fig. \ref{trajopt_vis} and summarized for all of the systems in Table \ref{trajopt_results}. 
	
	\emph{Box:} Inputs are optimized to move a stationary rigid body that is resting on a flat surface (Fig. \ref{gradient_compare}) to a goal location that is either to the right or up in the air $1$ meter. The planning horizon is $1$ second and the controls are initialized with zeros. Dojo uses a time step $h = 0.1$, whereas MuJoCo uses $h = 0.01$ to prevent significant contact violations with the floor. MuJoCo fails in the scenario with the goal in the air, while Dojo succeeds at both tasks.

	\emph{Hopper:} The hopping robot \cite{raibert1989dynamically} with $m = 3$ controls and $n = 14$ degrees-of-freedom is tasked with moving to a target pose over $1$ second. Similar, although not identical, models and costs are used. Dojo uses a time step $h = 0.05$ whereas MuJoCo uses $h = 0.01$. The hopper is initialized with controls that maintain its standing configuration. Quadratic costs are used to penalize control effort and perform cost shaping on an intermediate state in the air and the goal pose. The optimizer typically finds a single-hop motion.
	
	\emph{Quadruped:} The Unitree A1 with $m = 12$ controls and $n = 36$ degrees-of-freedom is tasked with moving to a goal location over a planning horizon $T = 41$ with time step $h = 0.05$. Controls are initialized to compensate for gravity and there are costs on tracking a target kinematic gait and control inputs. The optimizer finds a dynamically feasible motion that closely tracks the kinematic plan (Fig.~\ref{trajopt_vis}).
 
	Overall, we find that final results from both engines are similar. However, importantly, MuJoCo is enforcing soft contact whereas Dojo simulates hard contact. Dojo's gradients are computed with $\kappa = 3e{-}4$. Further, for systems with contact, MuJoCo requires a time step $h = 0.01$ for successful optimization, whereas Dojo succeeds with $h = 0.05$.

        \begin{figure}[t]
		\begin{center}
            \includegraphics[width=1.0\columnwidth]{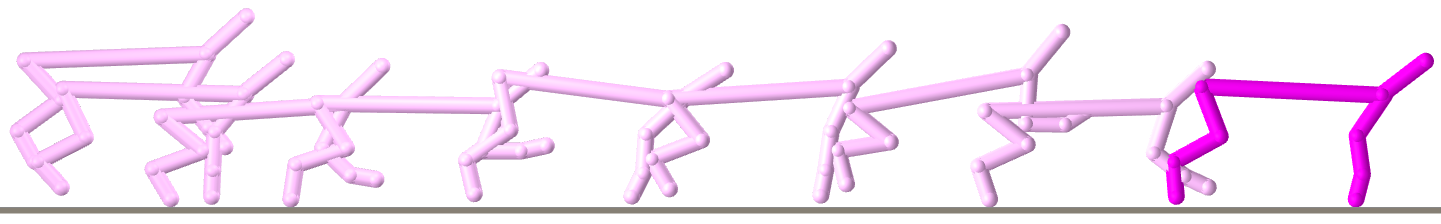}
            \includegraphics[width=1.0\columnwidth]{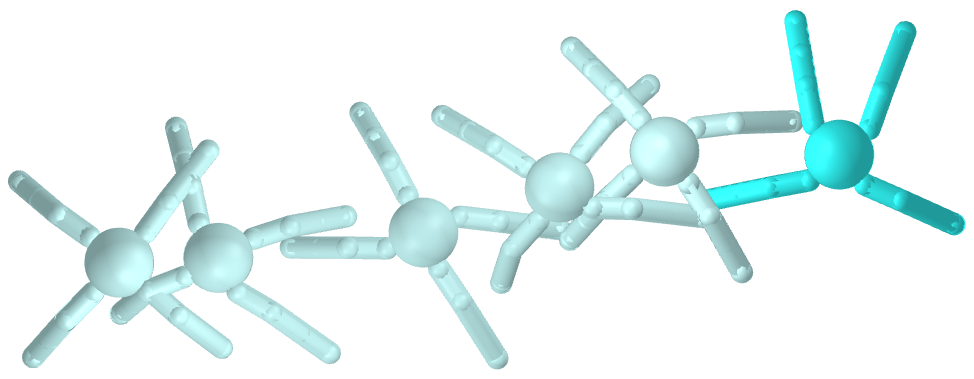}
		\end{center}
		\caption{Learned policy rollouts for half-cheetah (top) and ant (bottom). Time progresses left to right.}
		\label{rl_vis}
	\end{figure}
 
	\subsection{Policy Optimization}
	
	Gym-like environments \cite{brockman2016openai, duan2016benchmarking}: ant and half-cheetah are implemented in Dojo and we train static linear policies for locomotion. As a baseline, we employ Augmented Random Search (ARS) \cite{mania2018simple}, a gradient-free approach coupling random search with a number of simple heuristics. For comparison, we train the same policies using augmented gradient search (AGS) which replaces the stochastic-gradient estimation of ARS with Dojo's implicit gradients. Policy rollouts are visualized in Fig. \ref{rl_vis} and results are summarized in Table \ref{rl_results}.
	
	\emph{Half-cheetah:}
	This planar system with $m = 6$ controls and $n = 18$ degrees-of-freedom is rewarded for forward velocity and penalized for control effort over a horizon $T = 80$ with time step $h = 0.05$. 
	
	\emph{Ant:}
	The system has $m = 8$ controls and $n = 28$ degrees-of-freedom and is rewarded for forward motion and maintaining a certain altitude and is penalized for control effort and contact over a horizon $T = 150$ with time step $h = 0.05$.
	
	First, we are able to successfully train policies using this simple learning algorithm in Dojo's hard contact environments. Second, MuJoCo requires smaller $h = 0.01$ time steps for stable simulation, whereas Dojo is stable with $h = 0.05$. Third, our initial results indicate that it is possible to train comparable polices in Dojo with 5 to 10 times less samples by utilizing implicit gradients compared to the gradient-free method.

	\subsection{System Identification} 
    \begin{table}[t]
        \centering
		\caption{Policy optimization results. Comparison of total reward, number of simulation-step and gradient evaluations for a collection of policies trained with Augmented Random Search (ARS) \cite{mania2018simple} and Augmented Gradient Search (AGS). \qc{For AGS, we test with both implicit gradients and randomized gradients. The results are averaged over the best 3 out of 5 runs with different random seeds. Optimizing with implicit gradients from Dojo reaches similar performance levels while being 5 to 10 times more sample efficient, while optimizing with randomized finite difference gradients is 3 to 5 times more sample efficient.}} 
        \begin{tabular}{cccc}
        \toprule
        \textbf{Environment}          & \textbf{Method} & \textbf{Reward} & \textbf{Simulation Evals}          \\ \midrule
        \multirow{3}{*}{Half-Cheetah} & ARS                      & 46 $\pm$ 24        & 3e{+}4                           \\
                                      & AGS (\mac{Dojo})         & 44 $\pm$ 24        & \textbf{5e{+}3}                   \\
                                      & AGS (random)             & 39 $\pm$ 27        & 8e{+}3                            \\ \midrule
        \multirow{3}{*}{Ant}          & ARS                      & 64 $\pm$ 15        & 2e{+}5                             \\
                                      & AGS (\mac{Dojo})         & 54 $\pm$ 28        & \textbf{2e{+}4}                     \\
                                      & AGS (random)             & 59 $\pm$ 24        & 4e{+}4                               \\ \bottomrule
        \end{tabular}
        \label{rl_results}
    \end{table}
    
	System identification is performed on an existing real-world dataset of trajectories collected by throwing a box on a table with different initial conditions \cite{pfrommer2021contactnets}. We learn a set of parameters $\theta = (c_{\mathrm{f}}, p^{(1)}, \dots, p^{(8)})$ that include the friction coefficient $c_{\mathrm{f}}$, and 3-dimensional vectors $p^{(i)}$ that represent the position of vertex $i$ of the box with respect to its center of mass.
	
	Each trajectory is decomposed into $T-2$ triplets of consecutive configurations: $Z = (z_{-}, z, z_{+})$, where $T$ is the number of time steps in the trajectory. Using the initial conditions $z_{-}, z$ from a tuple, and an estimate of the system's parameters $\theta$, Dojo performs one-step simulation to predict the next state, $\hat{z}_{+}$. Implicit gradients are utilized by a Gauss-Newton method to perform gradient-based learning of the system parameters.
	    
    The parameters are learned by minimizing the following loss: 
	\begin{align}
	    \mathcal{L}(\mathcal{D}, \theta) = \sum_{Z \in \mathcal{D}} L(Z, \theta) = \sum_{Z \in \mathcal{D}} \frac{1}{2} ||\mbox{\textbf{Dojo}}(z_{-}, z; \theta) - z_{+}||_W^2,
	\end{align}
	where $||\cdot||_W$ is a weighted norm, which aims to minimize the difference between the ground-truth trajectories and physics-engine predictions. We use gradients
	\begin{equation}
	    \frac{\partial L}{\partial \theta} = {\frac{\partial \mbox{\textbf{Dojo}}}{\partial \theta}}^T W \left(\mbox{\textbf{Dojo}}(z_{-}, z; \theta) - z_{+} \right),
	\end{equation}
	and approximate Hessians 
	\begin{align}
	    \frac{\partial^2 L}{\partial \theta^2} &\approx {\frac{\partial \mbox{\textbf{Dojo}}}{\partial \theta}}^T W \frac{\partial \mbox{\textbf{Dojo}}}{\partial \theta}.
	\end{align}
	Gradients are computed with $\kappa = 3e{-}4$.
	
\begin{figure}[t]
		\begin{center}
            \begin{tikzpicture}
                \draw (0, 0) node[inner sep=0] {\includegraphics[width=\linewidth]{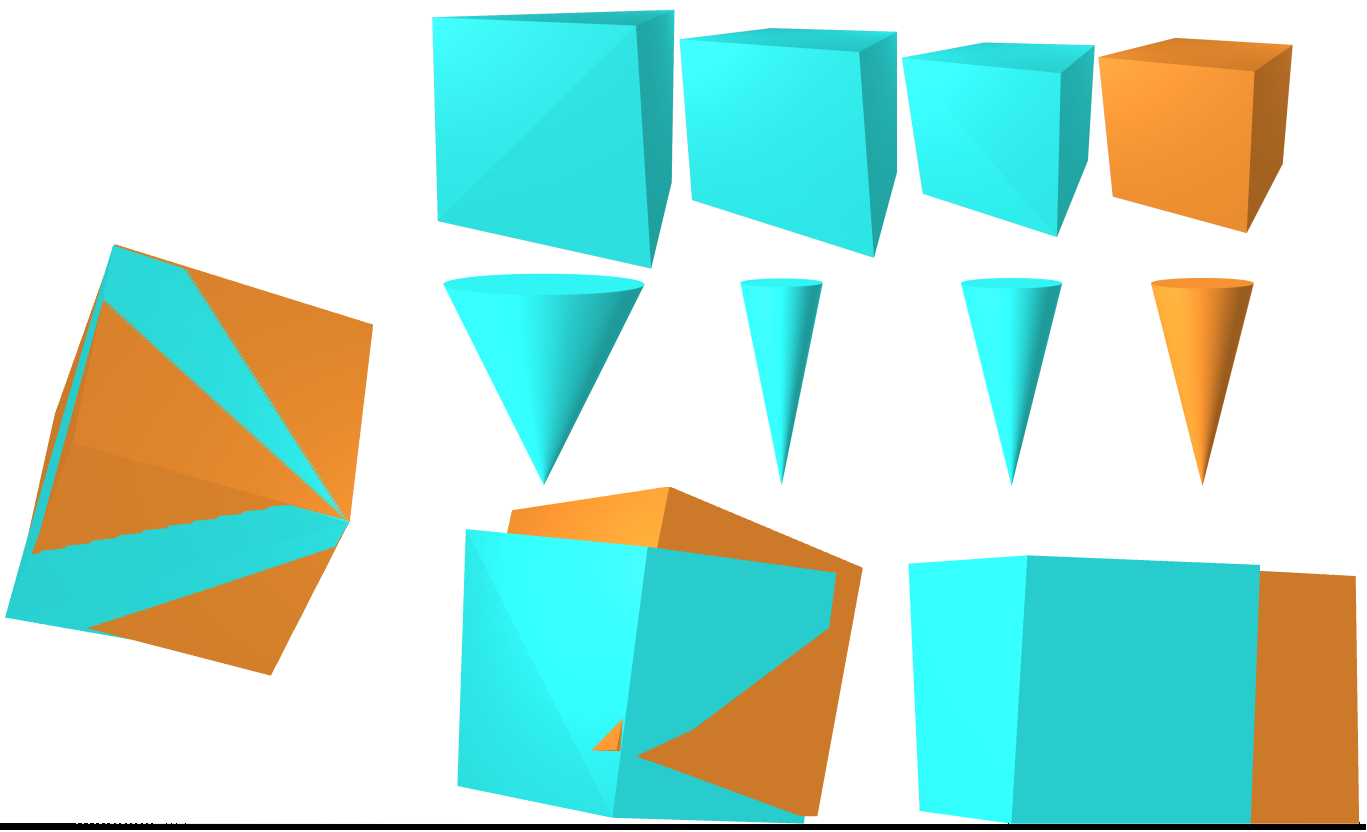}};
                \draw (-2.5, +2.9) node {Iteration:};
                \draw (-0.8, +2.9) node {(1)};
                \draw (+0.7, +2.9) node {(5)};
                \draw (+2.1, +2.9) node {(50)};
                \draw (+3.3, +2.9) node {truth};
            \end{tikzpicture}
		\end{center}
        
		\caption{System identification. Top right: Learning box geometry and friction cone to less than $5\%$ error. Bottom: Simulated trajectory of the box using the learned properties (blue) compared to ground truth (orange).}
		\label{real2sim}
	\end{figure}

	After training, the learned parameters are within $5\%$ of the true geometry and friction coefficient for the box from the dataset. We complete the real-to-sim transfer and simulate the learned system in Dojo, comparing it to the ground-truth dataset trajectories. Results are visualized in Fig. \ref{real2sim}.

    \section{Conclusion} \label{conclusion}
    Dojo is designed from physics- and optimization-first principles to enable better gradient-based optimization for planning, control, policy optimization, and system identification. 
    
        \subsection{Contributions}\label{contributions}
    The engine makes several advancements over previous state-of-the-art engines for robotics: First, the variational integrator enables stable simulation at low sample rates. Second, the contact model includes an improved friction model that eliminates artifacts like creep, particularly for sliding, and hard contact for impact is achieved to machine precision. This enables sim-to-real transfer for implementation on real robot hardware. The underlying primal-dual interior point solver, developed specifically for solving NCPs, is numerically robust and minimizes user hyperparameter tuning, while offering good performance across numerous systems, and handling cone and quaternion variables. Third, the engine efficiently returns implicit gradients whose smoothness through contact are tuned by the user to trade off gradient smoothness with simulation accuracy,  providing useful information through contact events. Fourth, in addition to building and providing Dojo as an open-source tool, the physics and optimization algorithms presented can be ported into existing simulation engines.
    
	\subsection{Limitations} \label{limitations}
    In terms of features, reliability, and wall-clock time, MuJoCo--the product of a decade of excellent software engineering--is impressive. As development of Dojo continues, we expect to make significant progress in all of these areas. However, fundamentally, Dojo's approach of solving an NCP with a primal-dual interior point method requires more computation per time step compared to existing simulators that use a soft-contact model (e.g. MuJoCo and Drake), but allows for accurate simulation with a lower sample rate, making wall-clock comparisons between the two simulators difficult. This is the fundamental trade-off Dojo makes for robotics applications: greater computational cost per time step for accurate physics and smooth gradients over fewer total time steps.
    
    Additionally, \mac{it should be acknowledged that Dojo's interior point solver is solving a nonconvex optimization problem at each simulation time step, which may have a danger of reaching poor local minima or not converging within the allotted time window.} Although in practice, we do not find this to be a problem, but for time- or safety-critical applications this should be a consideration. \qc{Moreover, the same is true in using Dojo's gradients for trajectory optimization, policy optimization, or system identification. These require solving inherently nonconvex optimization problems, which may converge to poor local solutions, or fail to converge, regardless of the smoothness or quality of Dojo's gradient information. Although smoother gradients may facilitate optimization for these problems, they do not fully resolve the complexities introduced by nonconvex optimization landscapes.}

	\subsection{Future Work} \label{future_work}
	
    A number of future improvements to Dojo are planned. First, Dojo currently implements simple collision detection (e.g., sphere-halfspace, sphere-sphere). Natural extensions include support for convex primitives and curved surfaces \mac{and triangular meshes}. Another improvement is adaptive time stepping. Similar to advanced numerical integrators for stiff systems, Dojo should take large time steps when possible and adaptively modify the time step in cases of numerical difficulties or physical inaccuracies. Finally, hardware-accelerator support for Dojo would potentially enable faster simulation and optimization.
    
    Perhaps the most important remaining question is whether the physics and optimization improvements from this work translate into better transfer of simulation results to successes on real-world robotic hardware. In this thrust, future work will explore the transfer of control policies trained in Dojo to hardware and deployment of the engine in model predictive control frameworks.
    
    In conclusion, we have presented a new physics engine, Dojo, specifically designed for robotics. This tool is the culmination of a number of improvements to the contact dynamics model and underlying optimization routines, aiming to advance state-of-the-art physics engines for robotics by improving physical accuracy and differentiability. 
	
	\section*{Acknowledgements}
    The authors would like to thank Suvansh Sanjeev for assistance with the Python interface. Toyota Research Institute provided funds to support this work. 
    
    \bibliographystyle{ieeetr}
    \bibliography{main}

    \appendices
    \section{Quaternion Algebra}
    \label{quaternion_algebra}
    In this section we introduce a set of conventions for notating standard quaternion operations, adopted from \cite{brudigam2020linear,jackson2021planning}, and employed in the rotational part of our variational integrator \eqref{rotational_integrator}.
    
    Quaternions are written as four-dimensional vectors:
    \begin{equation}
    	q = (s, v) = (s, v_1, v_2, v_3) \in \mathbf{H},
    \end{equation}
    where $s$ and $v$ are scalar and vector components, respectively. Dojo employs unit quaternions (i.e., $q^T q = 1$) to represent orientation, providing a mapping from the local body frame to a global inertial frame.
    
    Quaternion multiplication is represented using linear algebra (i.e., matrix-vector and matrix-matrix products). Left and right quaternion multiplication: 
    \begin{equation} 
    	q^a \cdot q^b 
    	= \begin{bmatrix}
    		s^a s^b - (v^a)^T v^b \\
    		s^a v^b + s^b v^a + v^a \times v^b
    	\end{bmatrix} 
    	= L(q^a)q^b = R(q^b) q^a ,
    \end{equation}
    where $\times$ is the standard vector cross product, is represented using the matrices:
    \begin{align}
    	L(q) &= \begin{bmatrix}
    		s & -v^T \\
    		v & s I_3 + \mathbf{skew}(v)
    	\end{bmatrix} \in \mathbf{R}^{4 \times 4},\\
    	R(q) &= \begin{bmatrix}
    		s & -v^T \\
    		v & s I_3 - \mathbf{skew}(v)
    	\end{bmatrix} \in \mathbf{R}^{4 \times 4},
    \end{align}
    where:
    \begin{equation}
    	\mathbf{skew}(x) = \begin{bmatrix}  
    		0 & -x_3 & x_2 \\
    		x_3 & 0 & -x_1 \\
    		-x_2 & x_1 & 0 
    	\end{bmatrix},
    \end{equation}
    is defined such that:
    \begin{equation}
    	\mathbf{skew}(x) y = x \times y ,
    \end{equation}
    and $I_3$ is a 3-dimensional identity matrix. The vector component of a quaternion:
    \begin{equation} 
    	v = V q,
    \end{equation}
    is extracted using the matrix:
    \begin{equation}
    	V = \begin{bmatrix}
    		\mathbf{0} & I_3
    	\end{bmatrix} \in \mathbf{R}^{3 \times 4},
    \end{equation}
    and quaternion conjugate: 
    \begin{equation} 
    	q^{\dagger} = \begin{bmatrix} s \\ -v \end{bmatrix} = T q,
    \end{equation}
    is computed using:
    \begin{equation}
    	T = \begin{bmatrix}
    		1 & \mathbf{0}^T \\
    		\mathbf{0} & -I_3
    	\end{bmatrix} \in \mathbf{R}^{4 \times 4}.
    \end{equation}

\end{document}